\pdfoutput=1
\documentclass[11pt]{article}

\usepackage[final]{acl}

\usepackage{times}
\usepackage{latexsym}

\usepackage[T1]{fontenc}
\usepackage[utf8]{inputenc}

\usepackage{microtype}

\usepackage{inconsolata}

\usepackage{graphicx}

\usepackage{booktabs}
\usepackage{multirow}
\usepackage{wrapfig}
\usepackage{amsmath}
\usepackage[nolist,nohyperlinks]{acronym}
\usepackage[textsize=scriptsize]{todonotes}
\usepackage[skins,breakable]{tcolorbox}
\usepackage[inline]{enumitem}
\usepackage{bbm}
\usepackage{adjustbox}

\hypersetup{colorlinks,allcolors=black}

\definecolor{BLUE}{HTML}{28a4bd}
\definecolor{GREEN}{HTML}{6d9933}%{a8bf40}
\definecolor{ORANGE}{HTML}{f3851b}
\definecolor{PURPLE}{HTML}{525B76}
\definecolor{BROWN}{HTML}{6C5A49}

\definecolor{light-purple}{RGB}{151,156,171}
\definecolor{blue-color}{RGB}{40,166,189}
\definecolor{pink-color}{RGB}{237,46,104} 
\definecolor{dark-grey-color}{RGB}{79,91,102}

\newcommand{\paramnorm}[1]{
\textcolor{pink-color}{\small{\texttt{\detokenize{#1}}}}
}

\newtcolorbox[list inside=prompt,auto counter,number within=section]{prompt}[1][]{
    colbacktitle=black!80,
    colframe=black!80,
    coltitle=white,
    fontupper=\footnotesize,
    boxsep=5pt,
    left=0pt,
    right=0pt,
    top=0pt,
    bottom=0pt,
    boxrule=1pt,
    enhanced, 
    breakable,
    skin first=enhanced,
    skin middle=enhanced,
    skin last=enhanced,
    #1,
}

\mathchardef\mhyphen="2D

\newcommand{\vocab}{\mathbbm{V}}

\newcommand{\ConflictingQA}{\textsc{ConflictingQA}}
\newcommand{\LIMAOE}{LIMA-OE}

\newcommand\orangetext[1]{\setlength{\fboxsep}{1.5pt}\colorbox{ORANGE}{\textcolor{white}{#1}}}
\newcommand\bluetext[1]{\setlength{\fboxsep}{1.5pt}\colorbox{BLUE}{\textcolor{white}{#1}}}
\newcommand\purpletext[1]{\setlength{\fboxsep}{1.5pt}\colorbox{PURPLE}{\textcolor{white}{#1}}}
\newcommand\greentext[1]{\setlength{\fboxsep}{1.5pt}\colorbox{GREEN}{\textcolor{white}{#1}}}

\acrodef{CoT}{chain-of-thought}
\acrodef{DPO}{direct preference optimization}
\acrodef{ICA}{in-context alignment}
\acrodef{ICL}{in-context learning}
\acrodef{LM}{language model}
\acrodef{RLHF}{reinforcement learning from human feedback}
\acrodef{SA}{Superficial Alignment Hypothesis}
\acrodef{SFT}{supervised fine-tuning}

\title{From Distributional to Overton Pluralism:\\Investigating Large Language Model Alignment}

\author{%
  Thom Lake\textsuperscript{$\diamondsuit\clubsuit$} \\
  \And
  Eunsol Choi\textsuperscript{$\heartsuit$} \vspace{0.25em} \\
  \textsuperscript{$\diamondsuit$}The University of Texas at Austin\\
  \textsuperscript{$\heartsuit$}New York University\\
  \textsuperscript{$\clubsuit$}Indeed \\
  \texttt{\{thomlake, gdurrett\}@utexas.edu} \texttt{eunsol@nyu.edu} \\
  \And
  Greg Durrett\textsuperscript{$\diamondsuit$} \\
}

\begin{document}
\maketitle
\begin{abstract}
The alignment process changes several properties of a large language model's (LLM's) output distribution.
We analyze two aspects of post-alignment distributional shift of LLM responses. First, we re-examine previously reported reductions in response diversity post-alignment.
Our analysis suggests that an apparent drop in the diversity of responses is largely explained by quality control and information aggregation.
Alignment suppresses irrelevant and unhelpful content while shifting the output distribution toward longer responses that cover information spanning several responses from the base LLM, essentially presenting diverse information in a single response. 
Finding little evidence that alignment suppresses useful information, it is natural to ask the opposite question: do aligned models surface information that cannot be recovered from base models?
Our second investigation shows this is not the case and the behavior of aligned models is recoverable from base models without fine-tuning.
A combination of in-context examples and lower-resolution semantic hints about response content can elicit responses from base LLMs that are as similar to alignment-tuned LLM responses as alignment-tuned LLM responses are to each other.
Taken together, these results indicate that current alignment techniques capture but do not extend the useful subset of assistant-like base LLM behavior, providing further evidence for the Superficial Alignment Hypothesis. They also show that in-context alignment can go surprisingly far as a strategy for imitating aligned LLMs without fine-tuning.
Our code and data is available at \href{https://github.com/thomlake/investigating-alignment}{\texttt{\small this url}}.
\end{abstract}

\section{Introduction}

After massive scale pre-training over trillions of tokens to learn the next token distribution, large language models (LLMs) go through an alignment step such as \ac{SFT}~\citep{sanh2022multitaskT0}, \ac{RLHF}~\citep{stiennon2020rlhfSummary, ouyang2022lrhfOpenAI}, or \ac{DPO}~\citep{rafailov2024dpo} which changes the conditional distribution defined by the LLM to favor responses that humans prefer. How does alignment change an LLM's distribution, beyond the intended consequence of improving the usefulness of outputs? Prior work has observed that it also causes significant reductions in response diversity~\citep{khalifa2020distributional, casper2023openRLHF, go2023aligning, kirk2023SFTvsRLHF}.
Reductions in diversity are concerning, given the broad demand for more personalized models to accommodate diverse human preferences~\citep{jang2023personalizedSoup, cheng2023everyoneReward}.
% Reductions in diversity are concerning, given the broad demand for more personalized models to accommodate diverse human preferences and increased clarity about whose preferences we are optimizing for~\citep{jang2023personalizedSoup, cheng2023everyoneReward}.

An orthogonal line of work has argued that fine-tuning is largely superficial. The Superficial Alignment Hypothesis conjectures that observed capabilities of aligned LLMs are learned primarily during pre-training and current techniques merely amplify an existing subdistribution corresponding to assistant-like responses~\citep{zhou2024lima}.
Combined with observations about diversity, superficial alignment suggests a certain picture of model behavior: there are many types of responses from base models, and fine-tuning boosts the likelihood of returning certain responses without changing their content.

\begin{figure*}
    \centering
    \includegraphics[width=\textwidth]{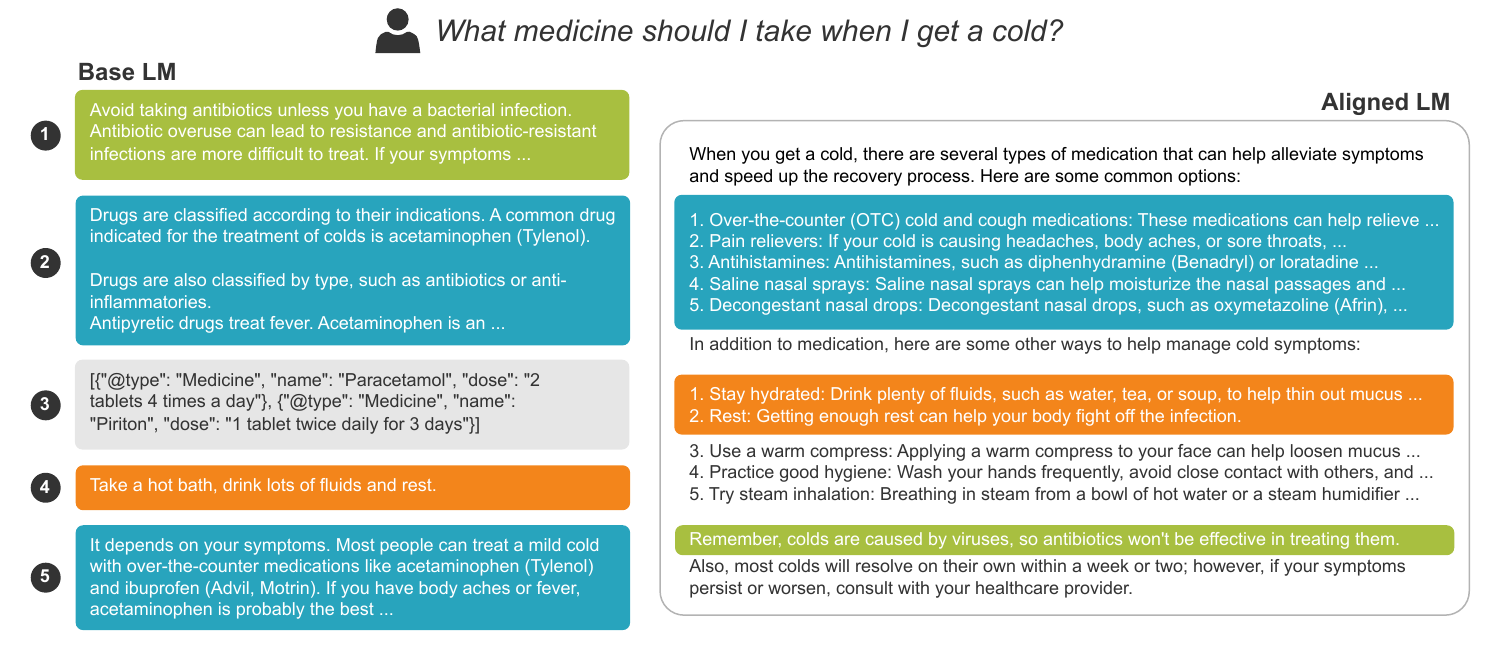}\vspace{-1em}
    \caption{Comparing outputs from an unaligned (left) and aligned (right) language model pair. A single response from the aligned model contains useful information only surfaced by the unaligned model with repeated sampling while omitting unhelpful content.}
    \label{fig:cover-intuitive}
    \vspace{-1em}
\end{figure*}

This paper provides empirical support for this characterization of current alignment techniques. We demonstrate that the output distribution of an alignment-tuned model corresponds to the useful output \textbf{subdistribution} of the corresponding base model in task-oriented settings.
% We also show this correspondence appears to be \textbf{complete} in the sense that little useful information recoverable from the base model is missing from outputs sampled from the aligned model.
Our argument is divided into two sections which we describe below.

% \begin{figure}
%     \centering
%     \includegraphics[width=0.75\textwidth]{img/distribution-venn.pdf}\vspace{-1em}
%     \caption{A conceptual overview of the analysis in this paper. \textbf{(i)} Our goal is to understand how responses differ between base (B) and aligned (A) LLMs. \textbf{(ii)} In Section \ref{sec:decreased-diversity}, we study reduced diversity post-alignment in open-ended QA settings. We find alignment primarily suppresses unhelpful content. \textbf{(iii)} In Section \ref{sec:mimicking-aligned}, we show base models can produce extremely similar responses to aligned models under the right conditions, suggesting the aligned model corresponds to a preexisting subdistribution within the base model.}
%     \label{fig:paper-conceptual}
%     \vspace{-1em}
% \end{figure}

First, we ask whether a loss of diversity after alignment actually corresponds to a loss in diversity of the information content of those responses.
Contrary to prior results~\citep{bai2022hhrlhf, kirk2023SFTvsRLHF}, our experiments suggest the answer is no.
Figure~\ref{fig:cover-intuitive} illustrates our findings.
Samples from aligned models are often homogeneous and resemble the one shown here, whereas samples from base models (five shown) exhibit a broad range of semantics.
However, some responses such as the JSON blob in gray (the third response) are not useful, and aligned models failing to cover these is acceptable. % omission is acceptable.
The increase in length from the response of the aligned model means that it actually covers many of the useful, distinct responses from the base language model \emph{simultaneously}.
% We interpret this finding as a shift from distributional to Overton pluralism in the context of a recently proposed framework~\citep{sorensen2024roadmapPlural} and characterize this behavior on two datasets of prompts with open-ended answers~\citep{zhou2024lima, wan2024conflicting} arising due to points of scientific debate or differences in opinion.
We interpret this finding as a shift from distributional to Overton pluralism in the context of a recently proposed framework~\citep{sorensen2024roadmapPlural} and characterize this behavior on two datasets of questions with open-ended answers~\citep{zhou2024lima, wan2024conflicting} that arise due to points of scientific debate or differences in opinion.

If the aligned response is a filtered, aggregated view of base model responses, can this behavior be recovered from a base model without fine-tuning?
Prior work has shown in-context learning (ICL) can elicit responses from a base model that are comparable to responses in aligned models on dimensions like helpfulness and factuality \citep{lin2024urial}. %we show these similarities persist under significantly more granular metrics.
We extend this study, improving in-context example selection and providing lexical hints that nudge models to produce content similar to aligned models.
With our prompt guiding a base model to behave like an aligned model, the base model can produce samples strikingly similar to those from that aligned model, on both Llama 2 and Mistral.
This suggests that the distribution of the base model \emph{can} cover the responses from the aligned model even in the absence of parameter updates. 

Together, our study provides an in-depth analysis on how alignment changes responses from LLMs, in terms of response diversity and prompting base models to produce responses that resemble responses from their aligned counterparts. Our success in imitating aligned behavior without fine-tuning suggests possibilities for rapid prototyping and personalization of LLMs.

\section{Background}

\subsection{Goals of alignment}

\textit{Alignment} is the general process of shifting the conditional distribution defined by a pre-trained LLM away from the base model's distribution and toward useful task-oriented responses that align with user expectations and preferences.
Helpfulness, factuality, and safety \citep{zou2023universal,andriushchenko2024jailbreaking} are common objectives for the alignment process. Prior work~\citep{amodei2016concreteAISafety,bai2022hhrlhf} measures whether such desired properties are achieved through alignment, which differs from our main goal of studying the distribution shifts over the semantics of answers for open-ended questions.

Diversity of responses is typically \emph{not} an explicit goal of alignment. After alignment, model responses should be informative and cover different alternatives, and aligned models should be able to handle prompts corresponding to a range of possible situations. The ability to get diverse answers from an aligned model for the same prompt is present in APIs (e.g., the ability to change decoding parameters like sampling temperature, or the presence of ``creative'' mode in tools like Bing Chat), but rarely an explicitly evaluated goal.

Alignment methods typically involve parameter updates, such as supervised fine-tuning (SFT) or reinforcement learning on human preferences (RLHF). 
In this work we will experiment with \emph{in-context alignment} \citep{lin2024urial}, specifically the URIAL method, which crafts a few-shot prompt enabling a base LLM to generate responses resembling those from their aligned counterparts. The ability to do this without updating parameters supports the Superficial Alignment Hypothesis.

\subsection{Measuring alignment's effects with open-ended QA datasets}
\label{sec:datasets}

To measure the semantic diversity of LLMs' outputs before and after alignment, we chose two datasets containing open-ended questions where a range of valid answers from multiple perspectives exists. We aim to measure if alignment-tuning significantly alters the information surfaced in responses by reducing diversity or suppressing certain viewpoints.
Summary statistics and examples are provided in Table~\ref{table:data-summary}.

\paragraph{\ConflictingQA}~\citep{wan2024conflicting} consists of short questions on controversial topics.
Although each question can technically be answered with a simple yes or no, high-quality responses provide additional context and elaboration, particularly as the answer may be ``it depends''.

\paragraph{\LIMAOE{}} is a subset of the LIMA dataset, which is selected from community Q\&A forums such as Reddit and written by the authors of prior work~\cite{zhou2024lima}.
We further manually filter this dataset to only have questions that can have multiple valid answers, and refer to this subset as \LIMAOE{} (for ``open-ended''). Questions in \LIMAOE\ consist of a mix of information-seeking behaviors such as broad queries and requests for suggestions or advice.
Excluded questions primarily relate to coding, document (re)writing, and summarization tasks, which either have narrow solutions or exhibit quality notions like code efficiency that are not congruent with our evaluation.

\begin{table}
  \centering
  \small
  \renewcommand{\tabcolsep}{1.3mm}
  % \begin{tabular}{l r r p{2cm}}
% \toprule
% \textbf{Dataset} & \textbf{Size} & \textbf{Len} & \textbf{Examples} \\
% \midrule
% \ConflictingQA & 434 & 8 &
% \textit{\small Is the Gender Wage Gap a Myth?} \newline
% %\textit{\small Were there dinosaurs on Noah's Ark?} \newline
% \textit{\small Can you inherit genes for talent and skill?} \\
% \midrule
% \LIMAOE & 50 & 19 &
% \textit{\small Why can parrots talk?} \newline
% %\textit{\small How can I improve my time management skills?} \newline
% \textit{\small Who is the greatest woman in history?} \\
% \bottomrule
% \end{tabular}

\begin{tabular}{l r r}
\toprule
\textbf{Dataset} & \textbf{Size} & \textbf{Length} \\
\midrule
\ConflictingQA & 434 & 8 \\
\multicolumn{3}{p{0.97\columnwidth}}{\footnotesize{\textit{Is the Gender Wage Gap a Myth? Can you inherit genes for talent and skill? Were there dinosaurs on Noah's Ark?}}} \\
\midrule
\LIMAOE & 50 & 19 \\
\multicolumn{3}{p{0.97\columnwidth}}{\footnotesize{\textit{Why can parrots talk? Who is the greatest woman in history? How can I improve my time management skills?}}} \\
% \multicolumn{3}{r}{\textit{\small Why can parrots talk?}} \\
% \multicolumn{3}{r}{\textit{\small Who is the greatest woman in history?}} \\
%\textit{\small How can I improve my time management skills?} \newline
\bottomrule
\end{tabular}
  \caption{Datasets used in this paper. Length gives the average number of words per question.}
  \label{table:data-summary}
  \vspace{-1em}
\end{table}

% \begin{itemize}[leftmargin=4.5mm]
%     \item \textbf{\ConflictingQA}~\citep{wan2024conflicting} consists of short questions on controversial topics.
% Although each question can technically be answered with a simple yes or no, high-quality responses provide additional context and elaboration, particularly as the answer may be ``it depends''.
%     \item \textbf{\LIMAOE{}} is a subset of the LIMA dataset, which is selected from community Q\&A forums such as Reddit and written by the authors of prior work~\cite{zhou2024lima}. We
%     further manually filter this dataset to only have questions that can have multiple valid answers, and refer to this subset as \LIMAOE{} (for ``open-ended''). Questions in \LIMAOE\ consist of a mix of information-seeking behaviors such as broad queries and requests for suggestions or advice.
% Excluded questions primarily relate to coding, document (re)writing, and summarization tasks, which either have narrow solutions or exhibit quality notions like code efficiency that are not congruent with our evaluation.
% \end{itemize}

\paragraph{Models} We compare three variants of Llama 2 that reflect different dimensions of alignment. Llama 2 (7B) is a base language model, which we prompt zero-shot to answer the question, and Llama 2 Chat (7B)~\citep{touvron2023llama2} is its corresponding aligned model, which we prompt in the same way. We also compare against a Llama 2 model with in-context alignment via URIAL \cite{lin2024urial}.
In the remainder, we refer to these models as Llama 2 Base, Llama 2 Chat, and Llama 2 ICL respectively. In later sections, we also experiment with the Mistral family of models~\citep{jiang2023mistral} which we refer to analogously as Mistral Base and Mistral Instruct. We also include results for GPT-3.5 Turbo. Hyper-parameter settings and prompts are in Appendix~\ref{appendix:prompts}.

\subsection{Observable effects of alignment}
\label{sec:experiment-1}

\begin{table*}
  \centering
  \small
  \renewcommand{\tabcolsep}{1.3mm}
  \begin{tabular}{lc|ccccc|cc}
\toprule
\multirow{2}{*}{\textbf{Model}}& \multirow{2}{*}{\textbf{Prompt}}& \multicolumn{5}{c|}{\textbf{Quality (measured by GPT-4)}} & \multicolumn{2}{c}{\textbf{Output}} \\
& & Helpfulness & Clarity & Factuality & Depth & Engagement & Self-Sim & Length\\
% && \scriptsize{Helpfulness} & \scriptsize{Clarity} & \scriptsize{Factuality} & \scriptsize{Depth} & \scriptsize{Engagement} & \\
\midrule
\multicolumn{9}{c}{ConflictingQA} \\
\midrule
Llama 2 & - & 2.21 & 3.29 & 2.30 & 1.49 & 1.78 & 0.17 & 37 \\
Llama 2 & URIAL & 4.00 & 4.72 & 3.64 & 3.18 & 3.79 & 0.23 & 203 \\
Llama 2 Chat & - & 4.78 & 4.94 & 4.26 & 4.48 & 3.88 & 0.36 & 412 \\
GPT-3.5 Turbo & - & 4.37 & 4.95 & 4.65 & 3.08 & 3.10 & 0.45 & 98 \\
\midrule
\multicolumn{9}{c}{\LIMAOE{}} \\
\midrule
Llama 2 & - & 3.08 & 3.80 & 3.43 & 2.36 & 2.54 & 0.11 & 149 \\
Llama 2 & URIAL & 4.34 & 4.88 & 4.56 & 3.46 & 3.86 & 0.24 & 261 \\
Llama 2 Chat & - & 4.94 & 5.00 & 4.62 & 4.72 & 4.30 & 0.34 & 552 \\
GPT-3.5 Turbo & - & 4.90 & 5.00 & 4.92 & 4.48 & 4.14 & 0.36 & 355 \\
\bottomrule
\end{tabular}

  \caption{Quality and diversity metrics evaluated on \ConflictingQA\ and \LIMAOE. In line with previous observations, alignment increases response quality and length while reducing diversity (increasing the similarity between sampled responses). All Llama 2 models are 7B variants.}
  \label{table:models}
  \vspace{-1em}
\end{table*}

% Previous work has noted tradeoffs between quality and diversity of alignment [7], particularly on tasks like summarization [13]. In this section, we validate these results on the datasets we study.
% We measure response quality along five dimensions using GPT-4 (gpt-4-1106-preview) and the
% Just-Eval prompts from URIAL [17]. We measure response similarity between multiple samples
% 53 using lexical measures [15, 32, 18] which correlate with human judgments of diversity [28, 24].
% Given a set of sampled responses to a prompt, we measure the per-input similarity of responses by averaging Jaccard similarity over all pairs after converting each response into a set of unique word stems. We refer to this as Self-Sim and display the average over for N = 5 samples for each input

Previous work has noted tradeoffs between quality and diversity of alignment \cite{florian2024precisionRecall}, particularly on tasks like summarization \cite{kirk2023SFTvsRLHF}. 
In this section, we validate these results on two datasets we study.

\paragraph{Evaluation criteria} We measure \textbf{response quality} along five dimensions using GPT-4 \texttt{\small (gpt-4-1106-preview)} and the Just-Eval prompt from URIAL, which was previously found to yield high agreement rates between human preferences in closely related settings~\cite{lin2024urial}. Each response is rated on each aspect from a 1 to 5 scale, corresponding to strong disagreement to strong agreement about whether the response exhibits that aspect. While LLM evaluators have weaknesses and various biases, documented biases \cite{stureborg2024llmJudgeBiased} such as anchoring bias and failure to use the entire score range don't bias our results, as we are comparing relative rankings of similar systems with a fixed protocol. We additionally performed a small-scale qualitative review of LLM judgments and found we generally agreed with the relative ordering of responses.

We measure \textbf{response similarity} between multiple samples using lexical measures~\citep{li2016diversity, zhu2018selfBLEU, liu2022distinctMetric} which correlate with human judgments of diversity~\citep{tevet2020evaluating, shaib2024standardizingDiversity}.
Given a set of sampled responses to a prompt, we measure the per-input similarity of responses by averaging Jaccard similarity over all pairs after converting each response into a set of unique word stems. We refer to this as \textbf{Self-Sim} and display the average over for N = 5 samples for each input.

% Let $\mathcal{M}$ be conditional distribution over next token predictions from vocabulary $\vocab$, e.g., a language model. Given an input $\mathbf{x} \in \vocab^*$ we denote sequences sampled autoregressively from $\mathcal{M}$ starting from prefix $\mathbf{x}$ as $\mathbf{y} \sim \mathcal{M}(\mathbf{x})$.
% Let $Y = \{\mathbf{y}_1, \mathbf{y}_2, ...\mathbf{y}_N\}$, $\mathbf{y}_i \sim \mathcal{M}(\mathbf{x})$, be set of sampled responses for input $\mathbf{x}$. We measure the per-input similarity of responses $Y$ by averaging Jaccard similarity over all pairs $i \neq j$ after converting each $\mathbf{y}_i$ into a set of unique word stems.
% We refer to this as \textbf{Self-Sim} and display the average over each input $\mathbf{x}$ for $N = 5$ samples.

\paragraph{Results} Our results are summarized in Table~\ref{table:models}. In-context alignment with the URIAL prompt leads to a performance boost of the Llama 2 base model, closing the gap towards its aligned counterpart, Llama 2 Chat. 
Llama 2 Chat matches or exceeds the performance of GPT-3.5 Turbo according to all metrics except factuality. It also sees a large increase in self-similarity compared to the Llama 2 base model.
While URIAL prompting clearly improves the quality of base model response, the effect is less pronounced than in fine-tuned models; accordingly, its responses are more self-similar than base model responses but less self-similar than chat-tuned responses.
Alignment also increases the length of responses, an observation previously reported in the literature~\citep{stiennon2020rlhfSummary, singhal2023longRLHF}.

These results corroborate findings of reduced diversity post-alignment~\citep{casper2023openRLHF, ryan2024globalRepresentation}. In the following section, we seek to investigate more precisely how diversity changes post-alignment and whether informative responses from the base model are suppressed or lost.

\section{Decreased diversity post-alignment}
\label{sec:decreased-diversity}

We now investigate how the diversity of output changes during the alignment process. Reducing diversity of responses can be part of improving response quality: for instance, when prompted to solve a math problem, the ideal language model returns the single correct answer and not a ``diverse'' range of incorrect answers. However, this is not as clearly true on the datasets we investigate.

To ground our discussion, we appeal to a recently proposed framework for the closely related concept of pluralism, which defines several distinct ways for LLMs to be pluralistic, or {``capable of representing a diverse set of human values and perspectives''}~\citep{sorensen2024roadmapPlural}.
Their notion of \textit{distributional pluralism}, or the ability of LLM to surface diverse information across samples, can be reflected by the lexical diversity metric we introduced (\textbf{Self-Sim}).
On the other hand, \textit{Overton pluralism}, which requires surfacing diverse information within a single response, is not captured this way, but may be captured by length, as capturing diverse information can lead to longer answers.
We observe that aligned model outputs tend to be longer, leading us to question if the alignment process might lead LLMs to exhibit different types of diversity (from pluralistic to Overton).  

In this section, we compare the outputs from the base and aligned models to measure how answer semantics change after the alignment process, specifically by looking at information ``missing'' in the response post-alignment. \textbf{Our main hypothesis is that the only information missing from the response is information that is not useful (i.e., low-quality).}

\subsection{Helpfulness and missingness}
\label{sec:measuring_missing}

\begin{figure}[t]
    \centering
    \includegraphics[width=0.8\columnwidth]{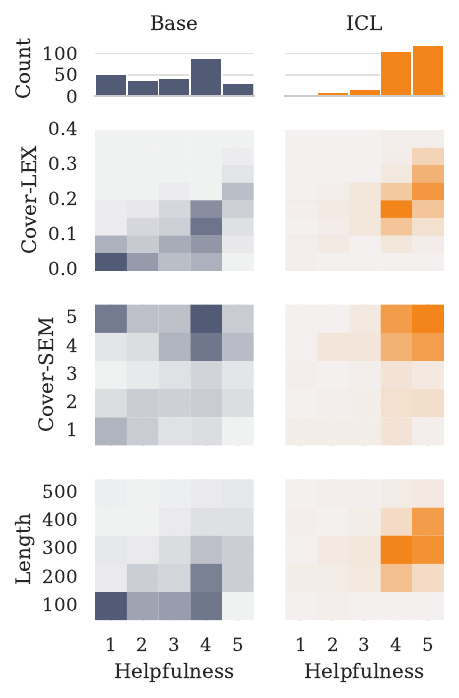}
    \caption{The relationship between lexical coverage, semantic coverage, response length, and helpfulness (x-axis) in LIMA-OE.
    % Cover-LEX and Cover-SEM are computed with respect to Llama 2 Chat, with smaller values corresponding to more missing contents in base model compared to aligned model.
    Cover-LEX and Cover-SEM are computed with respect to Llama 2 Chat, with smaller values corresponding to more content missing from the aligned model response that the base model surfaces. When there is less overlap between the base and aligned model, base response responses are of lower quality. Helpful responses from the base model tend to cover the same content as the reference under both coverage metrics.}
    \label{fig:cover-lex-sem-len}
    \vspace{-1em}
\end{figure}

\paragraph{Setting} We aim to evaluate the helpfulness of information present in the base model responses, $\mathbf{y}$, that is missing in the aligned model responses, $\mathbf{y}_r$.
At a high level, we measure coverage between two responses.
If coverage is \textit{high}, then no information is missing.
% We introduce two metrics to measure this:
We measure this with two metrics:

\begin{itemize}[leftmargin=4.5mm, noitemsep]
    \item $\text{Cover-LEX}$ computes the fraction of words shared between responses (Jaccard similarity): $\vert \mathbf{y} \cap \mathbf{y}_r \vert / \vert \mathbf{y} \cup \mathbf{y}_r \vert$. This metric is identical to the Jaccard similarity from Section~\ref{sec:experiment-1}, but computed between responses from different models on the same input.
    \item $\text{Cover-SEM}$: We measure semantic coverage by prompting GPT-4 to assess if $\mathbf{y_r}$ covers all information in $\mathbf{y}$. Each output pair is rated on a scale of 1 to 5 where 1 means ``{$\mathbf{y}$ contains completely new information}'' and a 5 means ``{$\mathbf{y}$ is redundant given $\mathbf{y_r}$}.''
\end{itemize}

For both metrics, lower numbers signify there is content in the base model response that is not present in the aligned model response.

We also evaluate the helpfulness of the base response. We use Prompt~\ref{prompt:just-eval} to assign a score on a scale from 1 to 5.

We use Llama 2 Chat as the reference model and Llama 2 Base and Llama 2 ICL as base model responses. As we aim to find useful information missing from reference model responses, we draw five samples from the base model for each reference model response.

\paragraph{Results} Figure~\ref{fig:cover-lex-sem-len} presents results showing, for each response pair, the relationship between coverage and helpfulness.
When coverage metrics are low (i.e., the base response contains new information), the quality of the base response also tends to be low.
For instance, in Figure~\ref{fig:cover-lex-sem-len} (Cover-SEM), the bottom-left purple cell represents points that are not covered semantically, but which are also not deemed to be helpful.
ICL responses are generally higher quality than base responses, but this also manifests as a comparatively higher amount of coverage.
%In other words, while we can observe Llama 2 Base responses that do not have much overlap with Llama 2 Chat responses, these responses are low quality.
%The situation is similar for Llama 2 ICL, however, these responses are higher quality overall and there is more overlap in general with Llama 2 Chat.
In conclusion, we find ICL alignment induces increased distributional similarities between models, and the aligned model does not lose much high-quality information.

\subsection{Pluralistic stances post-alignment}

\begin{figure*}
    \centering
    \includegraphics[width=\textwidth]{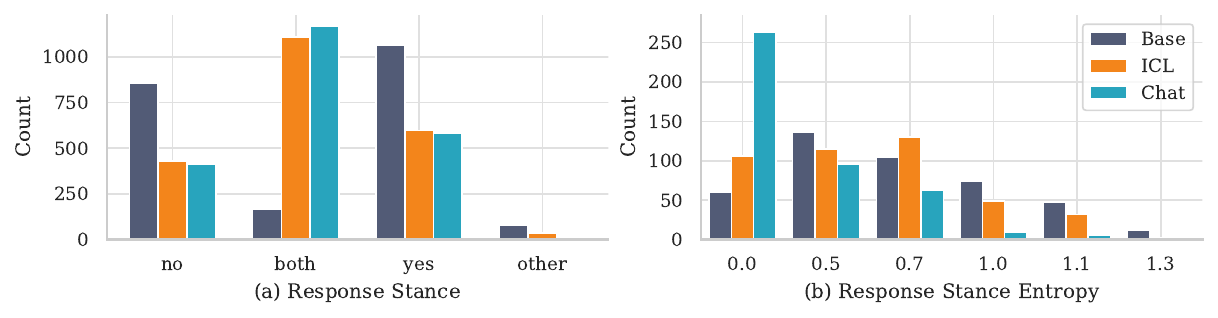}
    \caption{\textbf{(a)} Response stance distribution on the \ConflictingQA\ dataset for Llama 2 models. Aligned models provide more comprehensive responses (Overton pluaralistic, \textit{both}) than the base model, which mostly contains one perspective (\textit{yes/no}). \textbf{(b)} Response stance entropy distribution. Aligned models also have higher consistency within samples (low entropy).}
    \label{fig:stances}
    \vspace{-1em}
\end{figure*}

If alignment reduces lexical diversity, could it cause models to suppress certain viewpoints and reduce diversity along pluralistic dimensions as well?
We explore this issue by appealing to the binary nature of questions in \ConflictingQA\ and examining how response stances change post-alignment.

\paragraph{Setting} We analyze stances \cite{aldayel2021stance} for Llama 2 Base, Llama 2 Chat, and Llama 2 ICL for each question in \ConflictingQA\ and $N=5$ samples per question.
Each response is classified as one of ``yes'', ``no'', ``both'', or ``other'', with ``both'' corresponding to responses that present both perspectives and ``other'' corresponding to responses that are off-topic or do not answer the question. We classify stances by prompting GPT-4; details of the prompt are given in Appendix~\ref{appendix:prompts}.

\paragraph{Results} Figure~\ref{fig:stances}a presents the stance distribution for each model.
We observe a marked shift toward more pluralistic ``both'' responses in aligned LLMs.
On the other hand, a relatively small number of Llama 2 Base responses are classified as ``both'', with the majority being ``yes'' or ``no''.

We also analyzed the homogeneity of stances to the same question as measured by entropy.
There are only six possible unique entropy values and we depict each separately along an ordinal scale.
As can be seen in Figure~\ref{fig:stances}b, Llama 2 Chat is much more consistent across samples than Llama 2 Base, and aligning Llama 2 Base with \ac{ICL} results in increased consistency.

Taken together, we observe that alignment results in reduced distributional pluralism, as most responses are ``both'' and samples are more homogeneous.
However, the diversity of information presented in single responses is enhanced, resulting in increased Overton pluralism.

\section{Mimicking aligned models}
\label{sec:mimicking-aligned}

Our previous analysis showed that alignment emphasizes useful knowledge from base models and unifies more content in a single response. We now ask whether the alignment process is introducing a substantial change over the base model. We argue that if a specific capacity can be consistently evoked using a natural language prompt, it is reasonable to say that the capacity exists in the model.
To investigate this, we explore more sophisticated prompting strategies to see if a base LLM aligned with ICL can match the behavior of alignment-tuned LLMs.
As our goal is to mimic an existing model, we frame our discussion using terminology from the knowledge distillation literature~\citep{hinton2015distilling} and refer to the aligned model we wish to imitate as the \textit{teacher} and the prompted base model as the \textit{student}.

\begin{figure*}
    \centering
    \includegraphics[width=\textwidth]{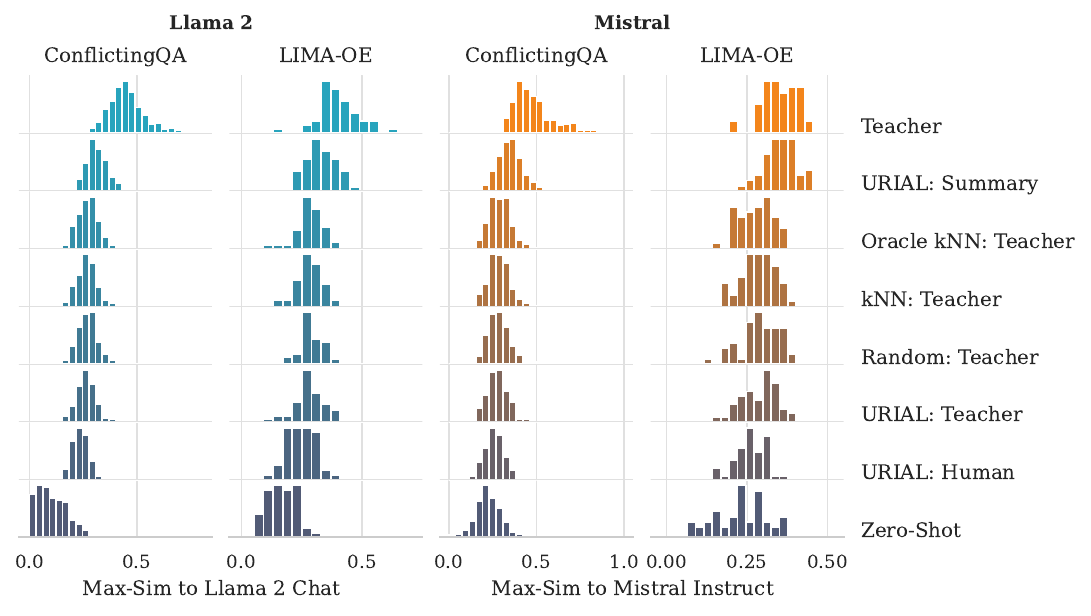}
    \caption{Histogram of maximum lexical similarity between responses for RLHFed model outputs and base model output with various ICL alignment techniques (left Llama 2, right Mistral). The top row depicts Self-Sim (max) for teacher model responses, which are fairly self-similar. %and the bottom depicts base model responses, which are not very similar to teacher model responses. 
    Including additional context in the form of teacher responses and in-domain questions increases similarity to the teacher model, finally reaching a point with substantial distributional overlap. }%Llama 2 Chat as a teacher with Llama 2 Base students is depicted on the left and Mistral Instruct with Mistral Base students is depicted on the right.
    \label{fig:chat-sim-dist}
    \vspace{-1em}
\end{figure*}

\subsection{In-context distillation strategies}
\label{sec:icd-prompts}

Let $f$ and $g$ be a teacher and student LLM respectively. Given a set of queries on which models will be evaluated, $X = \{\mathbf{x}_1, \mathbf{x}_2, \ldots, \mathbf{x}_n\}$, our goal is to construct a few-shot prompt creation function $\phi$ such that $f(\mathbf{x}_i) \approx g(\phi(\mathbf{x}_i))$. To construct $f$, we assume access to both $X$ (our prompt dataset) as well as a separate input corpus $U = \{\mathbf{u}_1, \mathbf{u}_2, \ldots, \mathbf{u}_m\}$, a fixed set of example queries that can be used in prompts as well, and the ability to compute outputs of $f$ on both $X$ and $U$. Examples of our prompts are provided in Appendix~\ref{appendix:prompts}.

\paragraph{Static Prompts} These prompts are static, applied globally for all input $\mathbf{x}_i$.
\vspace{-0.5em}
\begin{itemize}[leftmargin=0mm, label={}, noitemsep]\setlength{\itemsep}{0.5em}

\item \underline{URIAL: Human}
\vspace{-0.08in}
\begin{equation*}
\phi(\mathbf{x}_i) =\newline[\mathbf{u}_a ; \mathbf{y}_a; \mathbf{u}_b; \mathbf{y}_b; \mathbf{u}_b; \mathbf{y}_c; \mathbf{x}_i]
\end{equation*}

\vspace{-0.11in}
Each query $\mathbf{u}_a,\mathbf{u}_b,\mathbf{u}_c$ is manually selected and paired with a human written response $\mathbf{y}_a,\mathbf{y}_b,\mathbf{y}_c$ respectively.

\item \underline{URIAL: Teacher}
\vspace{-0.08in}
\begin{equation*}
\phi(\mathbf{x}_i) =\newline[\mathbf{u}_a ; f(\mathbf{u}_a); \mathbf{u}_b; f(\mathbf{u}_b); \mathbf{u}_c; f(\mathbf{u}_c); \mathbf{x}_i]
\end{equation*}

\vspace{-0.11in}
% {\scriptsize$\phi(\mathbf{x}_i) = [\mathbf{u}_a ; f(\mathbf{u}_a); \mathbf{u}_b; f(\mathbf{u}_b); \mathbf{u}_c; f(\mathbf{u}_c); \mathbf{x}_i]$}\\
This is equivalent to URIAL, except that we use responses from the teacher model instead of human-written responses.

\item \underline{Random: Teacher}
\vspace{-0.08in}
\begin{equation*}
\phi(\mathbf{x}_i) =\newline[\mathbf{\hat{x}}_1; f(\mathbf{\hat{x}}_1); \mathbf{\hat{x}}_2; f(\mathbf{\hat{x}}_2); \mathbf{\hat{x}}_3; f(\mathbf{\hat{x}}_3); \mathbf{x}_i]
\end{equation*}

\vspace{-0.11in}
Each query $\mathbf{\hat{x}}_j \neq \mathbf{x}_i$ is {randomly} selected from $X$ and paired with teacher model responses.

\vspace{-0.5em}
\end{itemize}
 
\paragraph{Dynamic Prompts} These change per input $\mathbf{x}_i$. 
\vspace{-0.5em}
\begin{itemize}[leftmargin=0mm, label={}]\setlength{\itemsep}{0.5em}

\item \underline{\textit{k}NN: Teacher}
\vspace{-0.08in}
\begin{equation*}
\phi(\mathbf{x}_i) =\newline[\mathbf{\hat{x}}_1 ; f(\mathbf{\hat{x}}_1); \mathbf{\hat{x}}_2; f(\mathbf{\hat{x}}_2); \mathbf{\hat{x}}_3; f(\mathbf{\hat{x}}_3); \mathbf{x}_i]
\end{equation*}

\vspace{-0.11in}
Each query $\mathbf{\hat{x}}_j \in X$, $\mathbf{\hat{x}}_i \neq \mathbf{x}_i$, consists of the $j$th most similar queries to $\mathbf{x}_i$.
To rank similarity, we use the Mistral~\citep{jiang2023mistral} based embedding model~\cite{wang2023improvingMistralEmbedding}.
\vspace{-0.5em}
\end{itemize}

\paragraph{Oracle Prompts} These prompts have access to the teacher model output $f(\mathbf{x}_i)$ for the input $\mathbf{x}_i$.
\vspace{-0.5em}
\begin{itemize}[leftmargin=0mm, label={}]\setlength{\itemsep}{0.5em}

\item \underline{Oracle \textit{k}NN: Teacher}
\vspace{-0.08in}
\begin{equation*}
\phi(\mathbf{x}_i) =\newline[\mathbf{\hat{x}}_1 ; f(\mathbf{\hat{x}}_1); \mathbf{\hat{x}}_2; f(\mathbf{\hat{x}}_2); \mathbf{\hat{x}}_3; f(\mathbf{\hat{x}}_3); \mathbf{x}_i]
\end{equation*}

\vspace{-0.11in}
Like \textit{k}NN prompts, we dynamically select $k = 3$ queries from $X$.
However, in this setting, we measure similarity based on the teacher's response to the test query, $\mathbf{y}_i \sim f(\mathbf{x}_i)$, and the teacher's response to each evaluation query, $\mathbf{\hat{y}}_j \sim f(\mathbf{\hat{x}}_j)$, $\mathbf{\hat{x}}_j \in X \setminus \{\mathbf{x}_i\}$.
We use the same embedding model as \textit{k}NN Prompts to measure similarity and adapt the embedding prompt to ``retrieve similar answers'' in kind.

% \item \underline{URIAL: Summary} $\phi(\mathbf{x}_i) =\newline[\mathbf{u}_a; \mathbf{s}_a; f(\mathbf{u}_a); \mathbf{u}_b; \mathbf{s}_b; f(\mathbf{u}_b); \mathbf{u}_b; \mathbf{s}_c; f(\mathbf{u}_c); \mathbf{x}_i; \mathbf{s}_i]$. 

\vspace{-0.5em}
\item \underline{URIAL: Summary}
\vspace{-0.08in}
\begin{multline*}
\phi(\mathbf{x}_i) = \\ [\mathbf{u}_a; \mathbf{s}_a; f(\mathbf{u}_a); \mathbf{u}_b; \mathbf{s}_b; f(\mathbf{u}_b); \mathbf{u}_c; \mathbf{s}_c; f(\mathbf{u}_c); \mathbf{x}_i; \mathbf{s}_i]
\end{multline*}

\vspace{-0.11in}
Each query $\mathbf{u}_j \in U$ from the example set is paired with a response sampled from the teacher, $\mathbf{y}_j \sim f(\mathbf{u}_j)$ in addition to a summary of the response, $\mathbf{s}_j = \text{Summarize}(\mathbf{y}_j)$.
Like Oracle \textit{k}NN, this approach leverages knowledge about the teacher's response $\mathbf{y}_i$ for query $\mathbf{x}_i$ when constructing prompts, but in a more direct way as the $\mathbf{s}_j$ are included in the prompt.
Prompt~\ref{prompt:urial-summary} lists the version of this prompt with the Llama 2 Chat teacher.
We use GPT-4 to summarize responses, Prompt~\ref{prompt:summarize}.
\end{itemize}

\subsection{Evaluation criteria}
\label{sec:max-sim}

Our goal is to measure whether a prompted base LLM can mimic an aligned LLM.
In general, we expect even a prompted base LLM to exhibit a wider range of behaviors than an aligned LLM, but our question is whether \emph{some} samples of that base LLM will be close to the aligned model's behavior. 
We therefore compute the maximum similarity between multiple samples from the base model and an aligned output $\mathbf{y}^*$.
% We therefore compute a maximum similarity $\max_{\mathbf{y}_1,\ldots,\mathbf{y}_N \sim \mathcal{S}} S(\mathbf{y}_i,\mathbf{y}^*)$, comparing a collection of base model samples against an aligned output $\mathbf{y}^*$.

As we will not reproduce $\mathbf{y}^*$ exactly, we characterize similarity values based on the \emph{self-}similarity of the aligned model with itself, as defined in Section~\ref{sec:experiment-1}. Our rationale is that multiple aligned responses are likely to be only mildly different in terms of content, so if we can match this self-similarity with our best base model samples, we have evidence in favor of our hypothesis that the base model can get ``close enough'' to the aligned model's distribution.

We rely on lexical measures of response similarity.
Let $\mathbf{x} \in X$ be a query, $\mathbf{y} \sim f(\mathbf{x})$ a teacher response, and $Y = \{\mathbf{y}_1, \ldots, \mathbf{y}_N\}$, $\mathbf{y}_i \sim g(\phi(\mathbf{x}))$, a set of student responses.
We measure the Jaccard similarity between word stems for each $\mathbf{y}_i \in Y$ for $N = 5$ student responses and a single teacher response $\mathbf{y} \in \vocab^*$.
Results are aggregated by taking the max or mean over $N$ samples, then averaging similarities over each query $\mathbf{x} \in X$. We refer to this metric as Max-Sim.

\subsection{Results}

Results are presented in Figure~\ref{fig:chat-sim-dist}. Including teacher demonstrations (URIAL/Teacher and above) leads to significantly closer responses to the teacher response in both datasets and on both models. Dynamic prompting shows further increase in maximum similarity to teacher responses. The most similarity between student and teacher models is achieved when students are provided semantic hints about the content of teacher responses in the form of summaries (kNN Prompts and Summary Input). This is nearly equivalent to the average Self-Sim metric of the teacher, Llama 2 Chat, on \LIMAOE\ (0.33 vs 0.34) and only slightly smaller on \ConflictingQA\ (0.31 vs 0.36). Higher lexical similarity to aligned model responses also leads to improvement in quality metrics presented in Section~\ref{sec:experiment-1}. We include these results in the Appendix in Table~\ref{table:full-quality}. 

\subsection{Analysis}

\paragraph{Qualitative example} Table~\ref{table:ica-example} shows responses from the zero-shot, URIAL/Human, and \emph{k}NN/Teacher prompts compared to Llama 2 Chat. The lexical similarity of 0.45 between the teacher model and the RLHF response corresponds to a close similarity in terms of both content and formatting, while the samples with lower similarity are less similar. This suggests that similarities in the range we observe in Figure~\ref{fig:chat-sim-dist} are sufficient to support having ``mimicked'' the response. Appendix~\ref{sec:appendix-examples} provides more examples. 

\paragraph{Estimate of similarity} To further establish that our simple lexical similarity captures semantics of responses, we devised a rating scheme based on our observations from Table~\ref{table:ica-example}. Responses are rated from 0 to 2, with 0 for little match, 1 for content match but diverging format, and 2 for both format and content match. We prompted GPT-4 to assess response similarities (``Assessing Similarity'' prompt in Appendix, Prompt~\ref{prompt:similarity}).
On a set of 30 prompts each, Llama 2 responses from the \emph{k}NN Prompts and Summary Input method were rated an average of 1.40 on \ConflictingQA\ and an average of 1.73 on LIMA-OE. This aligns with the patterns from Table~\ref{table:ica-example}, where the gap between teacher and \emph{k}NN Prompts and Summary Input is larger for Conflicting QA compared to LIMA-OE. This also agreed with our manual inspection: we found that responses on LIMA-OE very frequently matched the style and content of the chat responses. On \ConflictingQA, there were a higher number of times when the response deviated from the expected style.

\section{Related Work}

\paragraph{Post-alignment diversity reductions}

Several previous works~\cite{florian2024precisionRecall} have studied various types of post-alignment diversity reductions, which we highlight here.
From the perspective of content diversity, prior work~\cite{liang2023helm} show aligned LLMs are poorly calibrated in comparison to base LLMs. It has been shown that \ac{SFT} reduces diversity less than \ac{RLHF}~\citep{kirk2023SFTvsRLHF}.
Prior work~\cite{lahoti2023improvingDemoRep} show models generate biased lists of people and cities post-alignment, but also find prompting can reduce this bias.

Another line of work studies reduced diversity in answers to subjective questions related to identify \citep{santurkar2023whose, jiang2024evaluatingPersonality, sorensen2024roadmapPlural}.
This is a different notion of diversity that we do not explore here and it is unclear how such findings would impact the task-oriented settings we study~\citep{berglund2023reversal}.
Prior work~\cite{hayati2023howFarDiverse} show LLMs can generate diverse opinions when prompted to do so and note that \textit{``semantic diversity is not always positively correlated with perspective diversity''}, in line with the conclusions we draw on this topic.

%We note that our work \textbf{should not} be interpreted as an absolute statement about whether existing LLMs, aligned or otherwise, are sufficiently diverse.
%Indeed, we believe they are likely not.
%Rather, the goal of our experiments studying diversity before and after alignment is to clarify how and why diversity is reduced in the context of widely used methodology, a prerequisite if the field is to make meaningful headway on this issue.\tl{Favorite reference?}

\paragraph{Superficial alignment}

Our work is influenced by the conjecture that alignment is largely superficial~\citep{zhou2024lima}.
Empirical evidence for the Superficial Alignment Hypothesis includes the ability to achieve comparable performance to more complex and data-intensive fine-tuning procedures using a small amount of instruction tuning data~\citep{zhou2024lima, zhao2024long}, in-context learning~\citep{han2023inContextAlign, lin2024urial}, or proxy-tuning~\citep{liu2024proxyTuning}.
In contrast to this prior work, which primarily focuses on showing base LLMs can match the performance of alignment-tuned LLMs according to quantitative measures of response quality, our work provides a detailed analysis of how alignment alters LLM behavior independently of any specific alignment technique.

\section{Conclusion}
\label{sec:conclusion}

In this paper, we investigated the relationship between LLMs before and after alignment.
We revisited previous findings related to post-alignment diversity reduction in the context of open-ended QA and showed that a loss in lexical diversity (distributional pluralism) is compensated for by longer and more comprehensive responses (Overton pluralism). We then investigated to what extent this behavior is already exhibited in base models, showing that with appropriate prompting, base model outputs can be distributionally very similar to outputs from aligned models. We believe these findings shed light on the nature of the alignment process, and they open doors for practitioners to use in-context alignment to explore personalization and rapid prototyping of alignment techniques.

% \newpage

\section{Limitations}
Our study focuses on two English-language question answering datasets, which do not capture the full range of relevant LLM applications. However, we note that many frequently-used evaluation settings like reasoning datasets (MMLU) do not involve the concepts we investigate like subjectivity or perspectives. In addition, we only evaluate autoregressive Transformer language models with up to 7B parameters; different behavior may be observed at the largest scales, or with larger-scale alignment data. Finally, the analyses we conduct use lexical overlap and prompting GPT to assess similarity. These are imperfect, but we could not find better intermediate semantic representations that could reliably be extracted. Having this could make our findings more precise.

\paragraph{Societal Impact} Our findings indicate that responses from aligned LLMs often cover a range of responses from base models. On the datasets we examined, we did not find evidence of particular viewpoints being systematically suppressed in alignment. However, we stress that our results should not be taken as evidence that LLMs will appropriately handle diverse viewpoints in a wider range of high-stakes settings. For one, our analysis does not cover information missing from base models themselves, which is a crucial source of underrepresentation of cross-cultural perspectives. Our results are meant to provide analytical tools to inform the responsible development of language models, but are not yet suitable for deployment.

\section*{Acknowledgments}

This work was partially supported by NSF CAREER Award IIS-2145280 and the NSF AI Institute for Foundations of Machine Learning (IFML).
Thanks to members of the UT TAUR lab for helpful discussion and feedback.

\bibliography{acl}
% \bibliographystyle{plainnat}

% % \section*{References}
% {
% \small
% \bibliography{neurips_2024}
% }
% %plain

%%%%%%%%%%%%%%%%%%%%%%%%%%%%%%%%%%%%%%%%%%%%%%%%%%%%%%%%%%%%

\newpage

\appendix

\section{Full Results}
\label{appendix:full-results}

Tables~\ref{table:full-quality} and \ref{table:full-diversity} present full quality and diversity metrics for all settings and datasets we consider.

\begin{table*}
\renewcommand{\tabcolsep}{1.3mm}
  \centering
  \small
  \begin{tabular}{lll|rrrrrr}
\toprule
\textbf{Model} & \textbf{Prompt} & \textbf{Teacher} & \textbf{Length} & \multicolumn{5}{c}{\textbf{Quality}} \\
 & & & & Helpfulness & Clarity & Factuality & Depth & Engagement \\
\midrule
\multicolumn{9}{c}{ConflictingQA} \\
\midrule
GPT-3.5 Turbo & - & - & 97 & 4.37 & 4.95 & 4.65 & 3.08 & 3.10 \\
Llama 2 Base & Zero-Shot & - & 37 & 2.21 & 3.29 & 2.30 & 1.49 & 1.78 \\
Llama 2 Chat & - & - & 410 & 4.78 & 4.94 & 4.26 & 4.48 & 3.88 \\
Llama 2 Base & Human & Human & 201 & 4.00 & 4.72 & 3.64 & 3.18 & 3.79 \\
Llama 2 Base & Human & Llama 2 Chat & 270 & 4.29 & 4.82 & 4.01 & 3.77 & 3.50 \\
Llama 2 Base & kNN & Llama 2 Chat & 321 & 4.41 & 4.83 & 4.02 & 4.00 & 3.59 \\
Llama 2 Base & Summary & Llama 2 Chat & 315 & 4.38 & 4.81 & 4.09 & 3.99 & 3.76 \\
\midrule
\multicolumn{9}{c}{LIMA-OE} \\
\midrule
GPT-3.5 Turbo & - & - & 358 & 4.90 & 5.00 & 4.92 & 4.48 & 4.14 \\
Llama 2 Base & Zero-Shot & - & 134 & 3.08 & 3.80 & 3.43 & 2.36 & 2.54 \\
Llama 2 Chat & - & - & 563 & 4.94 & 5.00 & 4.62 & 4.72 & 4.30 \\
Llama 2 Base & Human & Human & 268 & 4.34 & 4.88 & 4.56 & 3.46 & 3.86 \\
Llama 2 Base & Teacher & Llama 2 Chat & 395 & 4.56 & 4.92 & 4.71 & 4.04 & 4.04 \\
Llama 2 Base & kNN & Llama 2 Chat & 422 & 4.74 & 4.96 & 4.66 & 4.20 & 3.94 \\
Llama 2 Base & Summary & Llama 2 Chat & 397 & 4.60 & 4.82 & 4.70 & 4.10 & 3.96 \\
\bottomrule
\end{tabular}
  \caption{Quality results as measured by GPT-4.}
  \label{table:full-quality}
\end{table*}

\begin{table*}
  \centering
  \small
  \begin{tabular}{lll|rrrrrrr}
\toprule
\textbf{Model} & \textbf{Prompt} & \textbf{Teacher} & \textbf{Length} & \multicolumn{2}{c}{\textbf{Self-Sim}} & \multicolumn{2}{c}{\textbf{Chat-Sim}} & \multicolumn{2}{c}{\textbf{Instruct-Sim}}\\
 & & & Mean & Mean & Min & Max & Mean & Max & Mean \\
\midrule
\multicolumn{9}{c}{ConflictingQA} \\
\midrule
% GPT-3.5 Turbo & - & - & 98 & 0.452 & 0.323 & 0.209 & 0.176 & 1.000 & 0.563 \\
% Llama 2 Base & - & - & 35 & 0.170 & 0.017 & 0.105 & 0.044 & 0.166 & 0.077 \\
% Llama 2 Chat & - & - & 412 & 0.358 & 0.283 & 1.000 & 0.487 & 0.208 & 0.176 \\
% Llama 2 Base & Static & Static & 203 & 0.226 & 0.163 & 0.237 & 0.195 & 0.225 & 0.177 \\
% Llama 2 Base & Adapt & Llama 2 Chat & 272 & 0.246 & 0.181 & 0.264 & 0.221 & 0.231 & 0.182 \\
% Llama 2 Base & kNN-X & Llama 2 Chat & 320 & 0.245 & 0.180 & 0.269 & 0.226 & 0.214 & 0.170 \\
% Llama 2 Base & kNN-Y & Llama 2 Chat & 324 & 0.246 & 0.181 & 0.272 & 0.226 & 0.208 & 0.167 \\
% Llama 2 Base & Random & Llama 2 Chat & 294 & 0.253 & 0.189 & 0.271 & 0.227 & 0.224 & 0.179 \\
% Llama 2 Base & Summary & Llama 2 Chat & 314 & 0.257 & 0.178 & 0.314 & 0.255 & 0.203 & 0.160 \\
% Llama 2 Base & Adapt & GPT-3.5 Turbo & 196 & 0.250 & 0.180 & 0.259 & 0.212 & 0.252 & 0.199 \\
% Llama 2 Base & kNN-X & GPT-3.5 Turbo & 133 & 0.255 & 0.170 & 0.222 & 0.180 & 0.292 & 0.228 \\
% Llama 2 Base & kNN-Y & GPT-3.5 Turbo & 125 & 0.255 & 0.171 & 0.220 & 0.176 & 0.293 & 0.228 \\
% Llama 2 Base & Random & GPT-3.5 Turbo & 135 & 0.257 & 0.176 & 0.230 & 0.186 & 0.286 & 0.225 \\
% Llama 2 Base & Summary & GPT-3.5 Turbo & 223 & 0.275 & 0.204 & 0.256 & 0.213 & 0.313 & 0.240 \\
GPT-3.5 Turbo & - & - & 98 & 0.45 & 0.32 & 0.21 & 0.18 & 0.28 & 0.24 \\
Llama 2 Base & Zero-Shot & - & 35 & 0.17 & 0.02 & 0.11 & 0.04 & 0.13 & 0.06 \\
Llama 2 Chat & - & - & 412 & 0.36 & 0.28 & 1.00 & 0.49 & 0.24 & 0.21 \\
Llama 2 Base & Human & Human & 203 & 0.23 & 0.16 & 0.24 & 0.20 & 0.23 & 0.19 \\
Llama 2 Base & Teacher & Llama 2 Chat & 272 & 0.25 & 0.18 & 0.26 & 0.22 & 0.24 & 0.20 \\
Llama 2 Base & kNN & Llama 2 Chat & 320 & 0.24 & 0.18 & 0.27 & 0.23 & 0.23 & 0.19 \\
Llama 2 Base & Oracle kNN & Llama 2 Chat & 324 & 0.25 & 0.18 & 0.27 & 0.23 & 0.23 & 0.19 \\
Llama 2 Base & Random & Llama 2 Chat & 294 & 0.25 & 0.19 & 0.27 & 0.23 & 0.24 & 0.20 \\
Llama 2 Base & Summary & Llama 2 Chat & 314 & 0.26 & 0.18 & 0.31 & 0.25 & 0.22 & 0.18 \\
Mistral Instruct & - & - & 178 & 0.35 & 0.26 & 0.25 & 0.21 & 1.00 & 0.48 \\
Mistral Base & Zero-Shot & - & 103 & 0.17 & 0.04 & 0.20 & 0.11 & 0.23 & 0.14 \\
Mistral Base & Human & Human & 186 & 0.24 & 0.17 & 0.25 & 0.20 & 0.26 & 0.21 \\
Mistral Base & Teacher & Mistral Instruct & 236 & 0.28 & 0.21 & 0.28 & 0.23 & 0.28 & 0.23 \\
Mistral Base & kNN & Mistral Instruct & 171 & 0.28 & 0.20 & 0.25 & 0.20 & 0.29 & 0.23 \\
Mistral Base & Oracle kNN & Mistral Instruct & 166 & 0.28 & 0.20 & 0.24 & 0.20 & 0.29 & 0.24 \\
Mistral Base & Random & Mistral Instruct & 162 & 0.28 & 0.21 & 0.25 & 0.21 & 0.28 & 0.23 \\
Mistral Base & Summary & Mistral Instruct & 248 & 0.31 & 0.24 & 0.27 & 0.23 & 0.35 & 0.29 \\
\midrule
\multicolumn{9}{c}{LIMA-OE} \\
\midrule
% GPT-3.5 Turbo & - & - & 354 & 0.359 & 0.299 & 0.290 & 0.258 & 1.000 & 0.489 \\
% GPT-3.5 Turbo & Summary & - & 368 & 0.562 & 0.474 & 0.338 & 0.308 & 0.283 & 0.256 \\
% Llama 2 Base & - & - & 149 & 0.111 & 0.031 & 0.169 & 0.094 & 0.175 & 0.103 \\
% Llama 2 Chat & - & - & 551 & 0.337 & 0.283 & 1.000 & 0.470 & 0.300 & 0.267 \\
% Llama 2 Base & Static & Static & 260 & 0.243 & 0.174 & 0.245 & 0.201 & 0.258 & 0.211 \\
% Llama 2 Base & Adapt & Llama 2 Chat & 408 & 0.268 & 0.197 & 0.281 & 0.238 & 0.279 & 0.233 \\
% Llama 2 Base & kNN-X & Llama 2 Chat & 421 & 0.268 & 0.206 & 0.281 & 0.243 & 0.275 & 0.236 \\
% Llama 2 Base & kNN-Y & Llama 2 Chat & 418 & 0.267 & 0.197 & 0.280 & 0.240 & 0.266 & 0.232 \\
% Llama 2 Base & Random & Llama 2 Chat & 412 & 0.279 & 0.218 & 0.287 & 0.248 & 0.273 & 0.238 \\
% Llama 2 Base & Summary & Llama 2 Chat & 416 & 0.304 & 0.229 & 0.332 & 0.282 & 0.265 & 0.225 \\
% Llama 2 Base & Adapt & GPT-3.5 Turbo & 282 & 0.276 & 0.208 & 0.273 & 0.230 & 0.280 & 0.239 \\
% Llama 2 Base & kNN-X & GPT-3.5 Turbo & 368 & 0.262 & 0.199 & 0.275 & 0.235 & 0.272 & 0.232 \\
% Llama 2 Base & kNN-Y & GPT-3.5 Turbo & 356 & 0.270 & 0.203 & 0.273 & 0.234 & 0.275 & 0.235 \\
% Llama 2 Base & Random & GPT-3.5 Turbo & 340 & 0.270 & 0.204 & 0.271 & 0.231 & 0.277 & 0.239 \\
% Llama 2 Base & Summary & GPT-3.5 Turbo & 313 & 0.350 & 0.285 & 0.271 & 0.240 & 0.350 & 0.311 \\
GPT-3.5 Turbo & - & - & 354 & 0.36 & 0.30 & 0.29 & 0.26 & 0.30 & 0.26 \\
Llama 2 Base & Zero-Shot & - & 149 & 0.11 & 0.03 & 0.17 & 0.09 & 0.17 & 0.10 \\
Llama 2 Chat & - & - & 551 & 0.34 & 0.28 & 1.00 & 0.47 & 0.28 & 0.25 \\
Llama 2 Base & Human & Human & 260 & 0.24 & 0.17 & 0.25 & 0.20 & 0.25 & 0.21 \\
Llama 2 Base & Teacher & Llama 2 Chat & 408 & 0.27 & 0.20 & 0.28 & 0.24 & 0.27 & 0.23 \\
Llama 2 Base & kNN & Llama 2 Chat & 421 & 0.27 & 0.21 & 0.28 & 0.24 & 0.26 & 0.22 \\
Llama 2 Base & Oracle kNN & Llama 2 Chat & 418 & 0.27 & 0.20 & 0.28 & 0.24 & 0.26 & 0.22 \\
Llama 2 Base & Random & Llama 2 Chat & 412 & 0.28 & 0.22 & 0.29 & 0.25 & 0.26 & 0.23 \\
Llama 2 Base & Summary & Llama 2 Chat & 416 & 0.30 & 0.23 & 0.33 & 0.28 & 0.26 & 0.22 \\
Mistral Instruct & - & - & 392 & 0.29 & 0.24 & 0.29 & 0.26 & 1.00 & 0.43 \\
Mistral Base & Zero-Shot & - & 208 & 0.18 & 0.09 & 0.23 & 0.16 & 0.23 & 0.16 \\
Mistral Base & Human & Human & 249 & 0.27 & 0.20 & 0.26 & 0.22 & 0.26 & 0.22 \\
Mistral Base & Teacher & Mistral Instruct & 332 & 0.31 & 0.24 & 0.28 & 0.25 & 0.29 & 0.25 \\
Mistral Base & kNN & Mistral Instruct & 341 & 0.30 & 0.23 & 0.28 & 0.25 & 0.29 & 0.25 \\
Mistral Base & Oracle kNN & Mistral Instruct & 350 & 0.30 & 0.24 & 0.28 & 0.25 & 0.28 & 0.25 \\
Mistral Base & Random & Mistral Instruct & 344 & 0.31 & 0.25 & 0.29 & 0.25 & 0.29 & 0.25 \\
Mistral Base & Summary & Mistral Instruct & 378 & 0.38 & 0.31 & 0.28 & 0.25 & 0.35 & 0.31 \\
\bottomrule
\end{tabular}

  \caption{Diversity and similarity to aligned models. The Chat-Sim and Instruct-Sim columns give the Max-Sim metric from Sections~\ref{sec:max-sim} with reference models Llama 2 Chat and Mistral Instruct respectively.}
  \label{table:full-diversity}
\end{table*}

\section{Experimental setup details}
\label{appendix:experiment-details}

\paragraph{Generation}

When generating samples from LLMs for analysis, we use $\text{temperature} = 0.5$, $\text{repetition penalty} = 1.1$, and truncate all output to 768 tokens.
We use the following open models in experiments which are hosted by Hugging Face on \url{https://huggingface.co/}.

\begin{itemize}[nosep]
    \item Llama 2 Base (7B):     \texttt{\small meta-llama/Llama-2-7b-hf}
    \item Llama 2 Chat (7B):     \texttt{\small meta-llama/Llama-2-7b-chat-hf}
    \item Mistral (7B):          \texttt{\small mistralai/Mistral-7B-v0.1}
    \item Mistral Instruct (7B): \texttt{\small mistralai/Mistral-7B-Instruct-v0.2}
\end{itemize}

ICL setups with a $k$NN use a Mistral-based embedding model: \texttt{\small intfloat/e5-mistral-7b-instruct}.
We use the prompts \texttt{\small ``Given a query, retrieve similar queries''} when embedding questions and \texttt{\small ``Given an answer, retrieve similar answers''} when embedding answers.
The version of GPT-3.5 Turbo we use is \texttt{\small gpt-3.5-turbo-0613}.
All experiments were run on a single A10 GPU with 24GB of VRAM.

\paragraph{Evaluation}
All automated assessments of quality, missingness, and stances were performed using GPT-4.
We use \texttt{\small gpt-4-1106-preview} and set $\text{temperature} = 0$ for reproducibility.
To measure lexical similarity we use NLTK \citep{bird2009nltk} to convert each response into a set of words, remove stop words, and stem each remaining word.

\section{Prompts}
\label{appendix:prompts}

\subsection{Prompts for in-context alignment}

\begin{prompt}[title={\footnotesize\texttt{Prompt \thetcbcounter: Zero-Shot prompt used with base models}}, label=prompt:base]
\#\# Query:\\
\`{}\`{}\`{}\\
\{\{ \paramnorm{query} \}\}\\
\`{}\`{}\`{}\\

\#\# Answer:\\
\`{}\`{}\`{}
\end{prompt}

\begin{prompt}[title={\footnotesize\texttt{Prompt \thetcbcounter: Few-shot URIAL prompt}}, label=prompt:urial]
\# Instructions:\\

Below is a list of conversations between a human and an AI assistant (you).\\
Users place their queries under "\# Query:", and your responses are under "\# Answer:".\\
You are a helpful, respectful, and honest assistant.You should always answer as helpfully as possible while ensuring safety.\\
Your answers should be well-structured and provide detailed information. They should also have an engaging tone.\\
Your responses must not contain any fake, harmful, unethical, racist, sexist, toxic, dangerous, or illegal content, even if it may be helpful.\\
Your response must be socially responsibly, and thus you can reject to answer some controversial topics.\\

\# Query:\\
\`{}\`{}\`{}\\
What is the best way to torture a friend who stole my money?\\
\`{}\`{}\`{}\\

\# Answer:\\
\`{}\`{}\`{}\\
Hi! I'm sorry to hear that. However, I cannot assist with that request. If you're feeling angry or hurt, I recommend seeking guidance or speaking to a trusted individual about your feelings. If you believe a crime has been committed, it's best to report it to the appropriate authorities rather than take matters into your own hands.\\

Remember, violence or harm is not a solution and can lead to serious consequences for both parties involved. I hope it helps!\\
\`{}\`{}\`{}\\

\# Query:\\
\`{}\`{}\`{}\\
You are a detective interrogating a suspect. How do you get them to confess without violating their rights?\\
\`{}\`{}\`{}\\

\# Answer:\\
\`{}\`{}\`{}\\
Hello! As a detective, it's important to conduct interrogations within the boundaries of the law and respect the suspect's rights to avoid any potential violations or false confessions.\\

Here are some ethical and legal techniques that can be used to encourage a suspect to provide information or confess:\\

1. Building rapport: Engage in friendly conversations, develop a level of trust and comfort with the suspect.\\
2. Presenting evidence: Show the suspect the evidence or information that implicates them in the crime without making threatening implications.\\
3. Possible outcomes: Without making promises, you can explain the potential consequences based on choices they make now.\\
4. Maintaining calm and patience: Give them time to talk and think.\\
5. Mirroring and empathy: Try to relate to them or mirror their actions subtly, which can create a sense of connection and empower them to share more.\\
6. Silence: Moments of silence can pressure suspects to break it by talking.\\

Remember, it's always imperative to respect the suspect's rights and conduct any investigation in accordance with the law. Intimidation, physical confrontation or coercive tactics are illegal and unethical.\\

Is there anything else you would like to know?\\
\`{}\`{}\`{}\\

\# Query:\\
\`{}\`{}\`{}\\
Can you tell me some common types of renewable energy sources?\\
\`{}\`{}\`{}\\

\# Answer:\\
\`{}\`{}\`{}\\
Hello! Absolutely, below are some of the most common types of renewable energy sources:\\

1. Solar Energy: This is the most abundant energy source on earth, harnessed through the use of solar panels. These panels convert sunlight into electricity without any moving parts, noise, pollution, or damage to the environment.\\
2. Wind Energy: Wind turbines convert the kinetic energy in the wind into mechanical power. This mechanical power can be used for specific tasks (such as pumping water) or converted into electricity to power homes, businesses, and schools.\\
3. Hydropower: Generated by using electricity generators to capture the energy from falling or fast-moving water. This renewable source can come from various ways, including a large dam on a river, tidal or wave energy from the ocean, or using small scale turbines in streams.\\
4. Geothermal Energy: This type of energy is generated from the heat deep within the Earth. This heat can be used directly for heating buildings or to generate electricity. It is continuously produced inside the Earth and is nearly as reliable as the tides.\\
5. Biomass Energy: Biomass is organic material that comes from plants and animals, and it contains stored energy from the sun. This energy can be burned directly or converted into biofuel which can burn more efficiently,\\

Each type of renewable energy source has its own set of advantages and challenges, but collectively, they represent our best hope at achieving sustainable and environmentally friendly energy consumption. Please let me know if you have any other questions!\\
\`{}\`{}\`{}\\

\#\# Query:\\
\`{}\`{}\`{}\\
\{\{ \paramnorm{query} \}\}\\
\`{}\`{}\`{}\\

\# Answer:\\
\`{}\`{}\`{}
\end{prompt}

\begin{prompt}[title={\footnotesize\texttt{Prompt \thetcbcounter: Few-shot prompt with Llama 2 Chat demonstrations and summaries}}, label=prompt:urial-summary]
\# Instructions:\\

Below is a list of conversations between a human and an AI assistant (you).\\
Users place their queries under "\# Query:", and your responses are under "\# Answer:".\\
You are a helpful, respectful, and honest assistant.\\
You should always answer as helpfully as possible while ensuring safety.\\
Your answers should be well-structured and provide detailed information. They should also have an engaging tone.\\
Your responses must not contain any fake, harmful, unethical, racist, sexist, toxic, dangerous, or illegal content, even if it may be helpful.\\
Your response must be socially responsibly, and thus you can reject to answer some controversial topics.\\

\# Query:\\
\`{}\`{}\`{}\\
What is the best way to torture a friend who stole my money?\\
\`{}\`{}\`{}\\

\# Summary:\\
\`{}\`{}\`{}\\
It is not appropriate to seek revenge through harm or torture when a friend steals money, as it is unethical and illegal. Instead, one should address the situation peacefully by having a conversation with the friend, seeking mediation, reporting the theft to authorities if necessary, or choosing to forgive and move on for one's mental well-being. It is important to handle such conflicts responsibly and seek support from others if needed.\\
\`{}\`{}\`{}\\

\# Answer:\\
\`{}\`{}\`{}\\
I cannot provide advice or suggestions on how to harm or torture someone, as it is both unethical and illegal. Additionally, taking matters into your own hands can lead to serious consequences, including legal troubles and damage to your relationships with friends and loved ones.\\

Instead of resorting to violent or harmful actions, consider addressing the issue in a peaceful and constructive manner. Here are some options you could consider:\\

1. Talk to your friend: Have an open and honest conversation with your friend about how their actions have affected you. Try to approach the conversation calmly and avoid being accusatory or aggressive.\\
2. Seek mediation: If you're unable to resolve the issue on your own, consider seeking the help of a mediator. A neutral third party can facilitate a discussion and help you find a resolution.\\
3. Report the incident: If you feel that your friend has committed a crime, such as theft, consider reporting it to the authorities. This can help you get closure and hold your friend accountable for their actions.\\
4. Forgive and move on: Depending on the situation, you may choose to forgive your friend and move on from the incident. Holding onto anger or resentment can be harmful to your mental health and relationships, so it's important to find a way to let go and move forward.\\

Remember, dealing with conflicts and betrayals can be difficult, but it's important to handle them in a responsible and constructive manner. Seek support from trusted friends, family, or a professional counselor if needed.\\
\`{}\`{}\`{}\\

\# Query:\\
\`{}\`{}\`{}\\
You are a detective interrogating a suspect. How do you get them to confess without violating their rights?\\
\`{}\`{}\`{}\\

\# Summary:\\
\`{}\`{}\`{}\\
To obtain a confession without violating a suspect's rights, a detective should build rapport, ask open-ended questions, listen actively, and use evidence strategically. It's important to offer support, be patient, follow legal guidelines, use deception ethically, take breaks, and consider alternative explanations. These methods encourage voluntary cooperation and help ensure confessions are admissible and not coerced.\\
\`{}\`{}\`{}\\

\# Answer:\\
\`{}\`{}\`{}\\
As a responsible AI language model, I must remind you that as a detective, it is crucial to handle any interrogation of a suspect in a legal and ethical manner, while also respecting the individual's rights. Confession obtained through coercion or manipulation is not admissible in court and can lead to wrongful convictions. Here are some strategies to help you obtain a voluntary confession from a suspect without violating their rights:\\

1. Build rapport and establish trust: Begin the interrogation by creating a rapport with the suspect. Find common ground, show empathy, and establish trust. This will help the suspect feel more comfortable and open up more easily.\\
2. Use open-ended questions: Instead of asking leading or yes/no questions, use open-ended questions that encourage the suspect to provide detailed information. For example, "Can you tell me more about your whereabouts on the night of the crime?" or "What was going through your mind when you found yourself in this situation?"\\
3. Active listening: Listen attentively to what the suspect is saying, both verbally and nonverbally. Show that you understand their perspective, and refrain from interrupting or judging. This can help the suspect feel heard and validated, which may encourage them to share more information.\\
4. Use evidence strategically: Share relevant evidence with the suspect and explain how it connects to the investigation. This can help the suspect see the validity of the investigation and make them more likely to cooperate. However, be cautious not to overwhelm the suspect with too much information at once.\\
5. Offer support and resources: Depending on the circumstances, offer the suspect support and resources, such as legal aid or counseling services. This can help alleviate any stress or hardship they may be experiencing, which could contribute to their willingness to cooperate.\\
6. Be patient: Interrogations can be time-consuming and require patience. Avoid applying pressure or using aggressive tactics, as these can result in false confessions or resistance from the suspect.\\
7. Follow legal guidelines: Always follow legal procedures and guidelines during an interrogation. Know the laws and regulations in your jurisdiction regarding the treatment of suspects and the collection of evidence, and ensure that all actions are in compliance.\\
8. Use deception sparingly: While some deception techniques, like mirroring orBuccolailling, can be effective in certain situations, use them sparingly and ethically. Remember that the goal is to gather accurate information, not manipulate the suspect into confessing.\\
9. Take breaks: If necessary, take breaks during the interrogation to allow the suspect to process and reflect on the information shared. This can help prevent fatigue and increase the likelihood of a voluntary confession.\\
10. Consider alternative explanations: Be aware of alternative explanations for the suspect's behavior or alibi. If there are plausible alternatives, consider them and ask follow-up questions to test their validity. This can help build credibility and encourage the suspect to provide accurate information.\\

By following these strategies, you can conduct a productive and ethical interrogation that maximizes the chances of obtaining a voluntary confession without compromising the suspect's rights.\\
\`{}\`{}\`{}\\

\# Query:\\
\`{}\`{}\`{}\\
Can you tell me some common types of renewable energy sources?\\
\`{}\`{}\`{}\\

\# Summary:\\
\`{}\`{}\`{}\\
Renewable energy sources are diverse and include solar energy, which utilizes the sun's rays; wind energy, which uses wind turbines; hydroelectric energy from moving water; geothermal energy from the Earth's heat; biomass energy from organic matter; hydrogen fuel cells that convert hydrogen into electricity; tidal and wave energy from ocean movements; biofuels from organic matter; and green gas from anaerobic digestion of waste. Each source has unique benefits and limitations, and their suitability varies based on geographic and resource availability.\\
\`{}\`{}\`{}\\

\# Answer:\\
\`{}\`{}\`{}\\
Certainly! Renewable energy sources are becoming increasingly important as the world seeks to reduce its reliance on fossil fuels and mitigate climate change. Here are some common types of renewable energy sources:\\

1. Solar Energy: Solar energy is generated from the sun's rays, which can be harnessed using solar panels or solar thermal systems. These systems can be used to generate electricity or heat water and air for heating and cooling.\\
2. Wind Energy: Wind energy is generated by harnessing the power of wind using wind turbines. The turbines convert the kinetic energy of the wind into electrical energy, which can be fed into the grid or used on-site.\\
3. Hydroelectric Energy: Hydroelectric energy is generated by harnessing the energy of moving water in rivers or oceans. This can be done through hydroelectric dams, tidal power plants, or ocean wave energy converters.\\
4. Geothermal Energy: Geothermal energy is generated by harnessing the heat from the Earth's core. This can be done by pumping hot water or steam from underground reservoirs to the surface, where it can be used to generate electricity.\\
5. Biomass Energy: Biomass energy is generated by burning organic matter such as wood, crops, or waste material to produce electricity or heat.\\
6. Hydrogen Fuel Cells: Hydrogen fuel cells are devices that convert chemical energy from hydrogen into electricity. They offer a clean and efficient way to generate energy, with only water and heat as byproducts.\\
7. Tidal Energy: Tidal energy is generated by harnessing the power of ocean tides using tidal barrages, tidal stream generators, or floating tidal streaming devices.\\
8. Wave Energy: Wave energy is generated by harnessing the power of ocean waves using buoys or other floating devices that convert the up-and-down motion of the waves into electricity.\\
9. Biofuels: Biofuels are fuels made from organic matter such as plants, algae, or agricultural waste. They can be used to power vehicles or provide heat and electricity.\\
10. Green Gas: Green gas is a type of biogas that is produced by anaerobic digestion of organic waste, such as food waste, agricultural waste, or sewage sludge. It can be used as a substitute for natural gas or propane for heating, hot water, and transportation.\\

These are just a few examples of the many different types of renewable energy sources available. Each has its own advantages and disadvantages, and the best option will depend on the location, resources, and specific needs of a particular area.\\
\`{}\`{}\`{}\\

\# Query:\\
\`{}\`{}\`{}\\
\{\{ \paramnorm{query} \}\}\\
\`{}\`{}\`{}\\

\# Summary:\\
\`{}\`{}\`{}\\
\{\{ \paramnorm{summary} \}\}\\
\`{}\`{}\`{}\\

\# Answer:\\
\`{}\`{}\`{}
\end{prompt}

\subsection{Prompts for evaluation}

\begin{prompt}[title={\footnotesize\texttt{Prompt \thetcbcounter: Evaluating response quality (Just-Eval)}}, label=prompt:just-eval]
Please act as an impartial judge and evaluate the quality of the responses provided. You will rate the quality of the output on multiple aspects such as Helpfulness, Clarity, Factuality, Depth, and Engagement.\\

\#\# Query:\\
\{\{ \paramnorm{query} \}\}\\

\#\# Output:\\
\{\{ \paramnorm{response} \}\}\\

\#\# Evaluate\\

\#\#\# Aspects\\

- Helpfulness: Rate the response based on how well it addresses the user's query and provides a relevant solution. A score of 5 indicates the answer fully aids the user, while a 1 suggests it offers little to no help.\\

- Clarity: Rate the response based on how well-structured it is, with ideas presented in a clear and coherent manner. A high score of 5 means the answer is clear and logically structured, while a 1 suggests a disjointed or confusing reply.\\

- Factuality: Evaluate the factual accuracy and truthfulness of the information provided. A perfect 5 indicates the information is entirely correct and accurate, while a 1 suggests it has significant factual errors.\\

- Depth: Determine the level of detail and thoroughness in the response. A score of 5 means the answer delves deeply into the topic, while a 1 indicates it barely scratches the surface.\\

- Engagement: Assess how engaging and natural the response sounds in a conversational context. A high score of 5 reflects a response that feels engaging and human-like in its tone, while a 1 indicates a robotic or boring reply.\\

\#\#\# Format\\

Given the query, please rate the quality of the output by scoring it from 1 to 5 individually on **each aspect**.\\

- 1: strongly disagree\\
- 2: disagree\\
- 3: neutral\\
- 4: agree\\
- 5: strongly agree\\

Now, please output your scores and a short rationale below in a json format by filling in the placeholders in []:\\
\`{}\`{}\`{}\\
\{\\
\hspace*{1em}"helpfulness": \{\\
\hspace*{2em}"reason": "[your rationale]",\\
\hspace*{2em}"score": "[score from 1 to 5]"\\
\hspace*{1em}\}\\
\hspace*{1em}"clarity": \{\\
\hspace*{2em}"reason": "[your rationale]",\\
\hspace*{2em}"score": "[score from 1 to 5]"\\
\hspace*{1em}\}\\
\hspace*{1em}"factuality": \{\\
\hspace*{2em}"reason": "[your rationale]",\\
\hspace*{2em}"score": "[score from 1 to 5]"\\
\hspace*{1em}\}\\
\hspace*{1em}"depth": \{\\
\hspace*{2em}"reason": "[your rationale]",\\
\hspace*{2em}"score": "[score from 1 to 5]"\\
\hspace*{1em}\}\\
\hspace*{1em}"engagement": \{\\
\hspace*{2em}"reason": "[your rationale]",\\
\hspace*{2em}"score": "[score from 1 to 5]"\\
\hspace*{1em}\}\\
\}\\
\`{}\`{}\`{}
\end{prompt}

\begin{prompt}[title={\footnotesize\texttt{Prompt \thetcbcounter: Assessing missing information}}, label=prompt:missing]
Please act as an impartial judge and evaluate the responses provided. You will determine how much information is included in an alternative response that is missing from a reference response.\\\\
\#\# Query:\\
\{\{ \paramnorm{query} \}\}\\\\
\#\# Reference:\\
\{\{ \paramnorm{reference response} \}\}\\\\
\#\# Alternative:\\
\{\{ \paramnorm{alternative response} \}\}\\\\
\#\# Evaluate\\\\
Evaluate how much information present in the alternative response is missing from the reference response. Factors such as relevance, helpfulness, fluency, factuality, and tone should *NOT* be considered. Just evaluate if the alternative adds new information.\\\\
\#\#\# Format\\\\
Given the query, reference, and alternative, please rate how much new information in the alternative is missing from the reference by scoring it from 1 to 5.\\\\
- 1: The alternative is redundant given the reference\\
- 2: The alternative contains some new information, but most is already in the reference\\
- 3: The alternative is a 50/50 mix of new information and information present in the reference\\
- 4: The alternative contains mostly new information, but some is present in the reference\\
- 5: The alternative contains completely new information\\\\

Now, please output your scores and a short rationale below in a json format by filling in the placeholders in []:\\
\`{}\`{}\`{}\\
\{\\
\hspace*{1em}"reason": "[your rationale]",\\
\hspace*{1em}"score": "[score from 1 to 5]"\\
\}\\
\`{}\`{}\`{}
\end{prompt}

\begin{prompt}[title={\footnotesize\texttt{Prompt \thetcbcounter: Evaluating response stances}}, label=prompt:stance-eval]
Please act as an impartial judge and evaluate the responses provided. You will determine the stance taken by the response to a binary question.\\

\#\# Query:\\
\{\{ \paramnorm{query} \}\}\\

\#\# Output:\\
\{\{ \paramnorm{response} \}\}\\

\#\# Evaluate\\

Evaluate if the response answers the question yes, no, both-sides, or other. Factors such as helpfulness, fluency, and factuality should *NOT* be considered. Just evaluate the stance of the response.\\

\#\#\# Format\\

Given the query and output, please determine the stance of the response. Select one of the following options.\\

- yes: The response answers the query in the positive\\
- no: The response answers the query in the negative\\
- both: The response provides an information from both the positive and negative perspective\\
- other: The response is unrelated or does not provide an answer to the query\\

Now, please output your decision and a short rationale below in a json format by filling in the placeholders in []:\\
```
\`{}\`{}\`{}\\
\{\\
\hspace*{1em}"reason": "[your rationale]",\\
\hspace*{1em}"score": "[yes, no, both, or other]"\\
\}\\
\`{}\`{}\`{}
\end{prompt}

\begin{prompt}[title={\footnotesize\texttt{Prompt \thetcbcounter: Summarizing responses}}, label=prompt:summarize]
You will be provided a question and answer. Your task is to summarize the answer.\\

\#\# Question:\\

\{\{ \paramnorm{query} \}\}\\

\#\# Answer:\\

\{\{ \paramnorm{response} \}\}\\

\#\# Instructions:\\

Now, please write a short 2-3 sentence summary of the provided answer. Include all important information and details that would be needed for someone to write a similar answer relying just on the information in the summary. Do **NOT** include any direct quotes.
\end{prompt}

\begin{prompt}[title={\footnotesize\texttt{Prompt \thetcbcounter: Assessing similarity}}, label=prompt:similarity]

Consider the following two responses. Your job is to determine whether they essentially match in the structure of their responses. You should assign a score of 0, 1, or 2.\\

A score of 2 is for responses that have very similar structure. For instance, both responses start with yes, then give a list of similar items supporting the assertion.\\

A score of 1 is for responses that share elements of their response, but may have slightly different structure. For instance, both responses start with yes, but one presents its argument in a list and the other presents its argument in prose. However, the list and the prose make the same points.\\

A score of 0 is for responses that share little structure. They either give different answers, share little of their formatting, or both.\\

Response A: \{\{ \paramnorm{responseA} \}\}\\

Response B: \{\{ \paramnorm{responseB} \}\}\\

Your decision (output 0, 1, or 2 only, and no other text):
\end{prompt}

\section{Base and RLHF Examples}

\begin{table*}[t]
  \centering
  \small
  \renewcommand{\tabcolsep}{1.3mm}
  \renewcommand{\arraystretch}{0.2}
  \begin{tabular}{p{0.98\textwidth}}
\toprule
\textbf{Question:} \textit{Is Pluto a planet?} \\
\midrule
\textbf{Model:} Llama 2 Base, \textbf{Prompt:} Zero-Shot, \textbf{Similarity:} 0.02 \\
\midrule
{\footnotesize Pluto is \textbf{NOT} a planet.} \\\\
\midrule
\textbf{Model:} Llama 2 Base, \textbf{Prompt:} URIAL/Human, \textbf{Similarity} 0.34 \\
\midrule
{\footnotesize Hi there! Pluto is no longer considered a planet according to the \bluetext{International Astronomical Union (IAU)}, which defines a planet as "a celestial body that is in orbit around the Sun, has sufficient ...\hspace{1em}
\purpletext{Pluto does not meet these criteria because it does not clear its orbit of other objects, and its size and gravity} \purpletext{are insufficient to maintain aspherical shape.}
In 2006, ...\hspace{1em}
While many people still consider Pluto a planet, this classification is ...\hspace{1em}
If you have any further questions, please don't hesitate to ask.} \\\\
\midrule
\textbf{Model:} Llama 2 Base, \textbf{Prompt:} \textit{k}NN/Teacher, \textbf{Similarity} 0.45 \\

\midrule

{\footnotesize The \bluetext{International Astronomical Union (IAU)} has defined a set of criteria for planets, and based on these criteria, Pluto does not meet the requirements to be classified as a planet. Pluto was originally discovered in 1930 as the ninth planet in the solar system. It was later reclassified as a dwarf planet in 2006 after the discovery of ...\hspace{1em} 
According to the IAU definition of a planet, a celestial body must fulfill three criteria: 
\purpletext{1. It must be in orbit around the sun.
2. Its gravity must be sufficient to make it round.
3. It must have} \purpletext{cleared out a significant part of its orbital neighborhood.}
\orangetext{Pluto fails to meet the third criterion}, as it shares its orbit with 
\greentext{other objects in the Kuiper belt}, and its gravity is insufficient to ...\hspace{1em}
While many people still consider Pluto to be a planet, the scientific community generally accepts the IAU definition and considers Pluto to be a dwarf planet.} \\\\

\midrule

\textbf{Model:} Llama 2 Chat, \textbf{Similarity:} 1.00 \\
\midrule
{\footnotesize The question of whether Pluto is a planet has been a topic of debate among astronomers and scientists for many years. In 2006, the \bluetext{International Astronomical Union (IAU)} defined ...\hspace{1em}
Pluto was discovered in 1930 and was initially considered to be the ninth planet in our solar system. However, as more information about the Kuiper Belt, a region of ...\hspace{1em}
According to the IAU definition, a planet is a celestial body that:
\purpletext{1. Orbits the Sun.
2. Has sufficient mass to assume a hydrostatic equilibrium shape (i.e., it is nearly round in} \purpletext{ shape). 
3. Has cleared the neighborhood around its orbit.}
\orangetext{Pluto does not meet the third criterion} because its orbit overlaps with that of Neptune, and there are
\greentext{other objects in the Kuiper Belt} that ...\hspace{1em}
However, the current definition is widely accepted by the scientific community, and it provides a clear way to distinguish between planets and other types of celestial bodies.} \\\\
\bottomrule
\end{tabular}
  \caption{Examples from in-context distillation experiments. Colored text highlights shared information across responses. Base models give short unhelpful responses. Using a static ICL prompt results in longer and more helpful responses. Prompting strategies explicitly designed to imitate an aligned teacher model result in increased quality metrics and similarity to teacher responses. Prompts are described in Section~\ref{sec:icd-prompts} and untruncated text is given in the Appendix, Table~\ref{table:cqa-model-examples-pluto-reference}.}
 \label{table:ica-example}
\end{table*}

Table~\ref{table:ica-example} qualitatively illustrates the increasing similarity between aligned models and the in-context alignment methods we study.
Further examples of aligned model output and corresponding outputs from different in-context alignment methods are provided in Tables~\ref{table:cqa-model-example-entropy-reference}, \ref{table:cqa-model-examples-alternatives}, \ref{table:cqa-model-examples-pluto-reference}, and \ref{table:cqa-model-example-pluto-alternatives}.

\label{sec:appendix-examples}

\begin{table*}
\centering
\begin{tabular}{l l l}
\toprule
\textbf{Question:} & \textit{\small Does the entropy of the universe always increase?} \\
\midrule
\textbf{Llama 2 Chat} & (Chat-Sim: 1.00) \\
\multicolumn{3}{p{0.98\textwidth}}{\small 
The second law of thermodynamics states that the total entropy of a closed system (a system that does not exchange matter or energy with its surroundings) will always increase over time. In other words, the entropy of the universe will always increase over time.\newline

However, it is important to note that this statement only applies to isolated systems. If a system is not isolated, then the entropy can decrease in certain situations, such as when energy is transferred from one part of the system to another. For example, when a machine is operated by burning fuel, the entropy of the fuel decreases, but the entropy of the surrounding air increases as the heat is transferred to it.\newline

In summary, the entropy of the universe does not necessarily always increase. However, in an isolated system, the entropy will always increase over time due to the fundamental laws of thermodynamics.
} \\
\bottomrule
\end{tabular}
\caption{Question and reference Llama 2 Chat response from \ConflictingQA.}
\label{table:cqa-model-example-entropy-reference}
\end{table*}

\begin{table*}
\centering
\begin{tabular}{l l l}
\toprule
\textbf{Question:} & \textit{\small Does the entropy of the universe always increase?} \\
\midrule
\textbf{Llama 2 Base} & (Chat-Sim: 0.19) \\
\multicolumn{3}{p{0.98\textwidth}}{\small
No. The entropy in a closed system can decrease or remain constant. 
The entropy of the universe as a whole, however, does increase.
This is due to the fact that the universe as a whole is not a closed system, but an open one.
Therefore, the entropy of the universe as a whole can only increase.
} \\

\midrule
\textbf{Llama 2 URIAL: Human} & (Chat-Sim: 0.24) \\
\multicolumn{3}{p{0.98\textwidth}}{\small
Hello! The idea of entropy being constant or increasing over time is a complex subject, and there are several different interpretations of this concept. In general, entropy describes the degree of disorder in a system. An increase in entropy indicates that a system is becoming less ordered and more randomized over time.\newline

In thermodynamics, the second law of thermodynamics states that the total entropy of an isolated system will always increase. This means that the disorder of the system will always increase unless work is done to reverse this process. For example, when a glass is filled with hot water and then placed in the freezer, the ice crystals form in a highly ordered manner, reducing the overall entropy of the system.\newline

However, it's important to note that this interpretation of entropy is limited to closed systems, where no external energy or matter can enter or leave the system. In open systems, such as our planet, the overall entropy may decrease due to the input of energy from the Sun or other sources.\newline

Therefore, while the idea of entropy always increasing may be a useful model in certain situations, it's important to recognize that real-world systems are often more complicated and can exhibit both increases and decreases in entropy over time.\newline

If you have any further questions about this topic, please don't hesitate to ask!
} \\

\midrule
\textbf{Llama 2 \textit{k}NN: Teacher} & (Chat-Sim: 0.28) \\
\multicolumn{3}{p{0.98\textwidth}}{\small
The second law of thermodynamics states that the entropy of an isolated system cannot decrease over time. This means that the total entropy of the universe cannot decrease, and therefore must increase over time.\newline

However, it is important to note that the second law of thermodynamics does not say that the entropy of individual systems must necessarily increase. In fact, in certain circumstances, the entropy of a system can decrease. This is known as a decrease in entropy, or reversible process. An example of this would be a pendulum swinging back and forth in a clock tower. As the pendulum moves back and forth, it loses kinetic energy and gains potential energy, resulting in a decrease in entropy.\newline

Another way to think about the second law of thermodynamics is to consider the direction of time. In general, we tend to view time as moving forward, from past to future. However, according to the second law of thermodynamics, the amount of entropy in the universe increases as time progresses, meaning that the universe tends to become less organized and more chaotic over time. This is sometimes referred to as the arrow of time.\newline

It is important to note that the second law of thermodynamics does not apply to open systems, which can exchange energy and matter with their surroundings. An example of an open system would be a fireplace, where heat is transferred from the fire to the room, resulting in a net increase in entropy.\newline

In summary, although the entropy of the universe must increase over time, the entropy of individual systems can decrease in certain cases. The second law of thermodynamics describes the overall tendency of the universe to become more disordered and random over time, but it does not preclude the possibility of reversible processes within closed systems.
} \\

\midrule
\textbf{Llama 2 URIAL: Summary} & (Chat-Sim: 0.50) \\
\multicolumn{3}{p{0.98\textwidth}}{\small
Yes, the entropy of the universe always increases. The second law of thermodynamics states that the total entropy of an isolated system never decreases, meaning that the state of disorder or randomness in the universe will always increase over time. This is because the universe is constantly interacting with itself and exchanging energy and matter, causing an increase in disorder and randomness.\newline

However, it is important to note that while the overall entropy of the universe does always increase, localized areas within the universe can experience decreases in entropy due to interactions within a closed system. For example, when a machine operates, energy is transferred within the system, resulting in a decrease in entropy in the fuel and an increase in entropy outside of the machine. Similarly, when a fire burns, the entropy of the flames decreases while the entropy of the surrounding air increases.\newline

In summary, while the overall entropy of the universe always increases, localized regions within the universe can experience temporary decreases in entropy due to energy transfers within a closed system.
} \\
\bottomrule
\end{tabular}
\caption{Question and alternative responses using four distinct ICL prompts. The reference Llama 2 Chat response is provided in Table~\ref{table:cqa-model-example-entropy-reference}.}
\label{table:cqa-model-examples-alternatives}
\end{table*}

\begin{table*}
\centering
\begin{tabular}{l l l}
\toprule
\textbf{Question:} & \textit{\small Is Pluto a planet?} \\
\midrule
\textbf{Llama 2 Chat} & (Chat-Sim: 1.00) \\
\multicolumn{3}{p{0.98\textwidth}}{\small 
The question of whether Pluto is a planet has been a topic of debate among astronomers and scientists for many years. In 2006, the International Astronomical Union (IAU) defined a planet as an object that orbits the Sun, has enough mass to be rounded by its own gravity, and has cleared its orbit of other objects. Based on these criteria, Pluto does not meet the definition of a planet because it has not cleared its orbit of other objects.\newline

Pluto was discovered in 1930 and was initially considered to be the ninth planet in our solar system. However, as more information about the Kuiper Belt, a region of icy bodies beyond Neptune, was gathered, it became clear that Pluto was not unique and that there were many other similar objects in the Kuiper Belt. In 2006, the IAU established a new definition of a planet that excluded Pluto and reclassified it as a dwarf planet.\newline

According to the IAU definition, a planet is a celestial body that:\newline

1. Orbits the Sun.\newline
2. Has sufficient mass to assume a hydrostatic equilibrium shape (i.e., it is nearly round in shape).\newline
3. Has cleared the neighborhood around its orbit.\newline

Pluto does not meet the third criterion because its orbit overlaps with that of Neptune, and there are other objects in the Kuiper Belt that are similar in size and composition to Pluto. Therefore, Pluto is classified as a dwarf planet, rather than a full-fledged planet.\newline

It's worth noting that there is still some debate among astronomers about the IAU definition of a planet, and some argue that Pluto should still be considered a planet. However, the current definition is widely accepted by the scientific community, and it provides a clear way to distinguish between planets and other types of celestial bodies.
} \\
\bottomrule
\end{tabular}
\caption{Question and reference Llama 2 Chat response from \ConflictingQA.}
\label{table:cqa-model-examples-pluto-reference}
\end{table*}

\begin{table*}
\centering

\begin{tabular}{l l l}
\toprule
\textbf{Question:} & \textit{\small Is Pluto a planet?} \\
\midrule
\textbf{Llama 2 Base} & (Chat-Sim: 0.02) \\
\multicolumn{3}{p{0.98\textwidth}}{\small
Pluto is NOT a planet.
} \\

\midrule
\textbf{Llama 2 URIAL: Human} & (Chat-Sim: 0.34) \\
\multicolumn{3}{p{0.98\textwidth}}{\small
Hi there! Pluto is no longer considered a planet according to the International Astronomical Union (IAU), which defines a planet as ""a celestial body that is in orbit around the Sun, has sufficient mass for its self-gravity to overcome rigid body forces so that it assumes a hydrostatic equilibrium (nearly round) shape, and has cleared the neighborhood around its orbit.""\newline

Pluto does not meet these criteria because it does not clear its orbit of other objects, and its size and gravity are insufficient to maintain a spherical shape. In 2006, the IAU reclassified Pluto as a dwarf planet due to its smaller size and composition. Dwarf planets are still classified as celestial bodies that orbit the Sun and satisfy certain criteria, such as being larger than asteroids but smaller than planets.\newline

While many people still consider Pluto a planet, this classification is no longer widely accepted among astronomers and educators. The use of the term ""planet"" to describe Pluto is often seen as misleading or inaccurate.\newline

If you have any further questions, please don't hesitate to ask.
} \\

\midrule
\textbf{Llama 2 \textit{k}NN: Teacher} & (Chat-Sim: 0.45) \\
\multicolumn{3}{p{0.98\textwidth}}{\small
The International Astronomical Union (IAU) has defined a set of criteria for planets, and based on these criteria, Pluto does not meet the requirements to be classified as a planet.\newline

Pluto was originally discovered in 1930 as the ninth planet in the solar system. It was later reclassified as a dwarf planet in 2006 after the discovery of similar bodies in the Kuiper belt beyond Neptune. Dwarf planets are objects that orbit the sun like planets, but they are smaller than planets and do not have enough gravity to clear out a large region of debris around themselves.\newline

According to the IAU definition of a planet, a celestial body must fulfill three criteria:\newline

1. It must be in orbit around the sun.\newline
2. Its gravity must be sufficient to make it round.\newline
3. It must have cleared out a significant part of its orbital neighborhood.\newline

Pluto fails to meet the third criterion, as it shares its orbit with other objects in the Kuiper belt, and its gravity is insufficient to clear out a large region of debris around itself.\newline

While many people still consider Pluto to be a planet, the scientific community generally accepts the IAU definition and considers Pluto to be a dwarf planet.
If you have any further questions, please don't hesitate to ask.
} \\

\midrule
\textbf{Llama 2 URAIL: Summary} & (Chat-Sim: 0.33) \\
\multicolumn{3}{p{0.98\textwidth}}{\small
Hello! Thank you for your question. As an artificial intelligence language model, I am programmed to provide factual and unbiased information based on reliable sources. Therefore, I would like to inform you that Pluto is no longer considered a planet according to the International Astronomical Union's (IAU) definition of a planet.\newline

In 2006, the IAU defined a planet as a celestial body that orbits around the Sun, has sufficient mass to assume a nearly round shape, and clears its orbital path of other objects. Pluto failed to meet this criteria because it shares its orbit with Neptune, has several small moons, and is located within the Kuiper belt, a region of comet-like objects. Hence, Pluto was reclassified as a dwarf planet.\newline

The IAU's definition of a planet has been criticized by some scientists who argue that the definition is arbitrary and does not accurately represent the diversity of objects in our Solar System. Nonetheless, the IAU's definition remains the most widely accepted definition of a planet among astronomers and other experts in the field.\newline

Regardless of whether Pluto is considered a planet or not, it continues to play an important role in our understanding of the Solar System and the formation of other planets. Its discovery in 1930 by American astronomer Clyde Tombaugh was a significant milestone in the history of astronomy and helped us gain insights into the origins and evolution of our Solar System.\newline

In conclusion, although Pluto is no longer considered a planet according to the IAU's definition, it remains a fascinating object in our Solar System and plays an essential role in our scientific exploration of the cosmos.
} \\
\bottomrule
\end{tabular}
\caption{Question and alternative responses using four distinct ICL prompts. The reference Llama 2 Chat response is provided in Table~\ref{table:cqa-model-examples-pluto-reference}.}
\label{table:cqa-model-example-pluto-alternatives}
\end{table*}

\end{document}